\documentclass{article}

\usepackage[english]{babel}

\usepackage[letterpaper,top=2cm,bottom=2cm,left=3cm,right=3cm,marginparwidth=1.75cm]{geometry}

\usepackage{amsmath}
\usepackage{graphicx}
\usepackage[colorlinks=true, allcolors=blue]{hyperref}
\usepackage{amssymb}
\usepackage{mathabx}
\usepackage{bbold}
\usepackage{multirow}
\usepackage{algorithm} 
\usepackage{algpseudocode}
\usepackage{diagbox}
\usepackage{subcaption}
\usepackage{rotating} 
\usepackage{enumitem}

\newcommand{\footremember}[2]{%
    \footnote{#2}
    \newcounter{#1}
    \setcounter{#1}{\value{footnote}}%
}
\newcommand{\footrecall}[1]{%
    \footnotemark[\value{#1}]%
}

\title{Variational Consensus Monte Carlo for Bayesian Mixture Models}
\author{Julie Fendler \footremember{BSU}{MRC Biostatistics Unit, University of Cambridge} \and  Francesca L. Crowe\footremember{Birmingham}{Institute of Applied Health Research, University of Birmingham} \and Tom Marshall\footrecall{Birmingham} \and Sylvia Richardson \footrecall{BSU}\and Paul D.W. Kirk\footrecall{BSU}}

\begin{document}
\maketitle

\begin{abstract}
Motivated by the privacy, sensitivity and sharing limitations of health data, we present a comprehensive pipeline for inference of Bayesian mixture models within a federated learning setting, i.e. when data cannot be fully shared or pooled across compute nodes. We adopt a Consensus Monte Carlo (CMC) approach, in which an MCMC algorithm is run independently within each data silo to estimate local posterior distributions, which are then aggregated to approximate the posterior over the full data. The variational CMC approach of Rabinovich, Angelino and Jordan (2015) \cite{rabinovich2015} frames the aggregation step as a variational inference problem, but their application to mixtures assumes the number of clusters and key mixture parameters to be known. Our main methodological contributions are: (i) an extension of variational CMC to over-fitted Bayesian mixture models that infer the number of clusters and all model parameters,  without requiring conjugacy; (ii) novel cluster-matching algorithms suitable for cross-silo settings in which not every cluster appears in each local dataset; (iii) a number of inference strategies for the aggregation step, matched to different federated learning constraints; and (iv) guidelines for choosing among these in practice. A comprehensive simulation study validates the framework and allows us to compare to state-of-the-art federated learning alternatives. Notably, we show that when the composition of local datasets reflects the underlying clustering structure in the data, our approach can recover  small clusters with greater accuracy than standard MCMC applied to the pooled data. We illustrate the framework on large-scale electronic health record data, identifying multi-morbidity patterns in a British geriatric population.
\end{abstract}

\section{Introduction}
According to Kairouz \textit{et al.} (2021) \cite{kairouz2021}, Federated Learning is a setting for machine learning in which data are split across multiple local computers (nodes). Model inference is coordinated by a central server (global node). None of the local nodes' data may be directly shared, exposed, or transferred to other local nodes or to the global node. In the federated learning literature, the dataset available to each local node is typically referred to as a {\em shard} of data and the union of all the shards forms the full dataset. Kairouz \textit{et al.} (2021) \cite{kairouz2021} distinguish three settings, although only two of these involve genuine restrictions on data sharing. The first, {\em data centre distributed learning}, artificially splits a single large dataset across nodes for computational reasons (to reduce memory or accelerate inference) and is best understood as distributed computing, rather than federated learning in the strict sense. {\em Cross-silo federated learning} refers to settings where the full dataset is naturally split into shards, e.g. according to a geographic pattern, meaning that data are geo-distributed. For example, in medical applications, data may be naturally siloed within the medical centres at which they are collected. Finally, in the {\em cross-device federated learning}, each shard contains only one observation of the full data, such as when data are split across personal mobile devices. Here we consider the cross-silo setting.\\
\newline
Recently, there has been growing attention toward FL for health data applications, evidenced by the surge in PubMed-indexed articles using the keyword ``federated learning" since 2019. Pfitzner, Steckhan and Arnrich (2021) \cite{PfitnerSteckhanArnrich2021} identify three main types of medical-related statistical task in which federated learning methods are used: 
\begin{enumerate}
    \item predicting an event from electronic health records (EHR), data from intensive care unit or omics data - these kinds of data can have a very large number of individuals and/or covariates leading to memory issues and are often pseudonymous, raising privacy concerns; 
    \item training models on image data, which are difficult to anonymise and therefore sensitive to share;
    \item training models on health data collected from smartphone or other wearable devices, which defines a cross-device federated learning setting.
\end{enumerate}
The authors emphasise that most of the literature regarding federated learning applications in healthcare focuses on supervised classification, with few studies addressing unsupervised clustering. The latter task, being commonly used to identify latent subgroups of patients in a given population, is the focus of our study. \\
\newline
In this work, we focus on unsupervised model-based clustering. We assume that the underlying generative mechanism of the data is a mixture model. Although the underlying model structure is assumed to be the same across shards, we do not assume that cluster prevalences are equal across shards, hence clusters may be observed in some shards but not others. As the number of clusters is unknown, we consider Bayesian over-fitted mixture models, i.e. rather than specifying the precise number of clusters in advance, we set the number of mixture components to be larger than the anticipated number of clusters, and rely on the posterior to indicate which components are non-empty \cite{RousseanMengersen2011, vanHavre2015}. To infer the model parameters within the Bayesian framework and in a federated learning setting, we adopt a {\em Consensus Monte Carlo} (CMC) approach. 
\subsection{Contributions and outline} %
Our main methodological contributions are:
\begin{enumerate}
\item An extension of variational CMC (VCMC) to over-fitted Bayesian mixture models, performing inference for the number of clusters and all model parameters without requiring conjugacy.
\item Novel cluster-matching algorithms suitable for cross-silo settings, in which not every cluster need appear in each shard.
\item A range of inference strategies for the aggregation step, matched to different federated learning constraints.
\item Guidelines for choosing among these in practice.
\end{enumerate}
Although we focus on mixtures of categorical distributions throughout, the framework requires only that MCMC inference is feasible in each shard and that a meaningful distance can be defined on the cluster-specific parameter space (needed for the matching step, Section \ref{sec:cluster_alignment}); it therefore extends to mixtures of more general distributions, including those outside the exponential family.\\
\newline
The remainder of the article is structured as follows. Section \ref{sec:backgrounds} provided a brief introduction to Consensus Monte Carlo and to the main existing methods to infer Bayesian mixture models in a federated learning setting. Section \ref{sec:StatModel} introduces our statistical model: Bayesian over-fitted mixtures of categorical distributions. Section \ref{sec:VCMC} presents the VCMC pipeline, including the matching algorithms and inference strategies. Section \ref{sec:simulation_study} describes a simulation study assessing the properties of our method and how it compares relative to state-of-the-art alternatives. Section \ref{sec:application_on_THINdata} applies the framework to large-scale electronic health record (EHR) data to identify patterns of multi-morbidity in a British geriatric population. We conclude with a discussion in Section \ref{sec:discussion}.
\section{Background}
\label{sec:backgrounds}
\noindent
The {\em Consensus Monte Carlo} approach of Scott {\em et al.} \cite{Scott2016} addresses Bayesian inference in distributed settings by running MCMC independently within each shard to produce local posterior samples, and then combining them to approximate the global posterior. The original proposal aggregates local draws by a weighted mean, with weights set to the inverse of the local-posterior variance-covariance matrices. This is simple and robust, but provides no guarantee that the resulting combination optimally approximates the global posterior.\\
\newline
Rabinovich, Angelino and Jordan (2015) \cite{rabinovich2015} refine this by treating the aggregation weights as the target of variational inference. The weights are optimised to minimise the Kullback-Leibler divergence  between the approximated and true global posterior distribution. The authors show that a relaxed version of the evidence lower bound (ELBO) is block-convex under mild conditions, making the variational problem tractable. They demonstrate that this approach leads to smaller errors in the first and second order moment approximations of the global posterior distribution compared to the methods of Scott {\em et al.} \cite{Scott2016} when applied to Gaussian mixture models. In their examples, the authors assume that the number of clusters, the mixture weights, and the cluster-specific covariance matrices are all known, and also implicitly assume that every cluster is present in each shard. In practice, none of these is likely to hold, hence relaxing these assumptions is one of the main contributions of the present work.\\
\newline
To our knowledge,  no other work has extended the Rabinovich, Angelino and Jordan VCMC framework to the more general settings we consider here.  The limited number of studies that have focused on Bayesian federated clustering instead adopt alternative approaches. We summarise four representative methods here.  SNOB \cite{Zuanetti2019} infers Dirichlet process Gaussian mixtures in two stages: local inference within each shard, followed by global inference under the constraint that locally-identified clusters cannot be split. Only local cluster sizes and sufficient statistics need to be shared to perform the second step. SIGN \cite{Ni2020} extends this approach to Pitman-Yor process mixture models and non-conjugate settings, with additional intermediate steps for scalability. However, in non-conjugate settings, the method requires access to the full dataset in its final step, which makes it unsuitable for cross-silo federated settings. DisCGS \cite{Khoufache2025} provides a detailed empirical study of distributed Dirichlet process Gaussian mixtures across diverse datasets, showing that these types of methods can perform accurate clustering. However, the goodness of the posterior density estimate is not assessed. Most closely related to our present work, FedMerDel \cite{Rao2025} provides a variational framework to infer over-fitted mixtures of categorical distributions in federated settings. Like the previous methods, it operates in two stages: first performing local inference within each shard, then merging locally-identified clusters to identify the global clustering structure. This method does not require data sharing at the global level other than the local estimates of the parameters of the posterior distributions. However, it uses a mean-field variational approximation and requires conjugate priors for tractable update equations.\\
\newline
\section{Statistical model}
\label{sec:StatModel}
Let $n\in \mathbb{N}^*$ denote the number of individuals in the full data. For $i$ in $\{1,...,n\}$, we observe $\boldsymbol{X}_i = (X^1_i, ..., X^Q_i)$, $Q \in \mathbb{N}^*$ categorical covariates (i.e., $\forall q \in [\![ 1, Q]\!], X^q_i \in \{1,...,K^q\}, K^q \in \mathbb{N}^*$).\\
\newline
We now assume that there exists an underlying clustering mechanism of the data. Conditional on the cluster memberships of the individuals, $(C_i)_{i \in [\![1,n]\!]}$, the observed covariates follow cluster-specific categorical distributions: 
$$\forall i \in [\![1, n]\!], \forall q \in [\![ 1, Q]\!], X_i^q|C_i \sim \mbox{Categorical}(p^q_{.,C_i})$$
$$C_i \sim \mbox{Categorical}(\pi)$$
with: \begin{itemize}
    \item $\forall c \in [\![1, C^+]\!], \forall q \in [\![ 1, Q]\!], p^q_{.,c} = (p^q_{1,c},...,p^q_{K^q,c}) \in [0,1]^{K^q} \mbox{ and }\sum_{k=1}^{K^q}p_{k,c}^q = 1$;
    \item $\forall c \in [\![1, C^+]\!], \pi = (\pi_1, ...,\pi_{C^+}) \in [0,1]^{C^+} \mbox{ and } \sum_{c=1}^{C^+} \pi_c = 1$;
    \item $C^+ \in \mathbb{N}, C^{+} \geq C =  \max_{i \in [\![1, n]\!]} C_i$.
\end{itemize}  
The above equations define a mixture of $C^{+}$ categorical distributions. We set $C^{+}$ to be large but finite, which defines an over-fitted mixture model \cite{RousseanMengersen2011}. The likelihood of the model can be written by integrating out the latent cluster membership variable: \begin{equation*}
    \mathbb{P}(\boldsymbol{X_i}|\pi, \boldsymbol{p}^1, ...,\boldsymbol{p}^Q) = \sum_{c = 1}^{C^+} \pi_{c} \prod_{q = 1}^{Q} p_{X_i^q, c}^q
\end{equation*} \\
where $\forall q \in [\![1,Q]\!], \boldsymbol{p}^q = \{p^q_{.,1}, ..., p^q_{.,C^+}\}$
\newline
Rousseau and Mengersen (2011) \cite{RousseanMengersen2011} show that under suitable prior distribution on the mixture weights, the posterior distribution asymptotically converges to a mixture model with $C$ non-empty clusters. In accordance with their recommendation and in order to achieve conjugacy, we define the following prior distributions: 
$$\pi \sim \mbox{Dirichlet}(\alpha)$$
$$\forall q, \forall c, p^q_{.,c} \sim \mbox{Dirichlet}(\beta^q) $$
with: \begin{itemize}
    \item $\alpha \in \mathbb{R}^{C^+}, \forall c \in [\![1, C^+]\!], \alpha_c >0 $;
    \item $\forall q \in [\![1,Q]\!], \beta^q \in \mathbb{R}^{K^q}, \forall k \in [\![1, K^q]\!], \beta^q_k >0 $.
\end{itemize} 
In practice, we set $\alpha$ and $\beta^q$ to 0.5 for all $q$ in $q \in [\![1,Q]\!]$. These values define non-informative Jeffreys prior distributions, consistent with the recommendation of Rousseau and Mengersen (2011) \cite{RousseanMengersen2011}. 

\section{Variational Consensus Monte Carlo (VCMC)}
\label{sec:VCMC}
\subsection{Consensus Monte Carlo}
We now introduce notation and describe the Consensus Monte Carlo framework. We define $\boldsymbol{\dot{X}}$ to be the full data. We say that $\boldsymbol{\dot{X}}$ is split into $S$ shards if there exists a partition, $(\boldsymbol{\dot{X_1}}, \boldsymbol{\dot{X_2}}, ..., \boldsymbol{\dot{X_S}})$, of $\boldsymbol{\dot{X}}$ of size $S$, where for all $s$ in $[\![1,S]\!]$ there exists $(i_1, ...,i_{n_s})$, $n_s$ distinct elements of $[\![1,n]\!]$ such that $\boldsymbol{\dot{X_s}} = (\boldsymbol{X}_{i_1}, \boldsymbol{X}_{i_2},  ..., \boldsymbol{X}_{i_{n_s}})$. \\
\newline
If the data points of $\boldsymbol{\dot{X}}$ are i.i.d., the global posterior distribution of the model parameters $p(\theta | \boldsymbol{\dot{X}} )$ (i.e. the posterior distribution estimated over the full data $\boldsymbol{\dot{X}}$) can be written 
$$p(\theta |\boldsymbol{\dot{X}}) = p(\theta) \prod_{s=1}^S \ell(\boldsymbol{\dot{X_s}}|\theta) =\prod_{s=1}^S \ell(\boldsymbol{\dot{X_s}}|\theta) p(\theta)^{\frac{1}{S}}$$ where $\ell(\boldsymbol{\dot{X_s}}|\theta)$ is the likelihood of the model evaluated with the data points in shard $s$ and $p$ the prior distribution of the model parameters $\theta$. \\
\newline
In the following, we will write $\tilde{p}_s(\theta) = p(\boldsymbol{\dot{X_s}}|\theta) p(\theta)^{\frac{1}{S}}$, the local posterior distribution inferred in shard $s, s\in [\![1,S]\!]$. $ p(\theta)^{\frac{1}{S}}$ is called the fractionated prior distribution.\\
\newline
{\em Consensus Monte Carlo} consists of two steps \cite{Scott2016}:
\begin{itemize}
    \item the apply step where a Bayesian sampling-based method is used to draw samples from each of the $S$ local posterior distributions;
    \item the aggregate step where the local posterior distributions are combined to infer the global posterior distribution.
\end{itemize}

\subsection{The apply step}
Samples from each local posterior distribution are drawn using an MCMC algorithm. This step can be performed using any Bayesian sampling-based method. We use a Gibbs sampler as, in our model defined in Section \ref{sec:StatModel}, conjugacy is achieved for all the marginal distributions. Under the fractionated prior, the marginal distributions of the parameters of a categorical mixture model in a shard $s$ are as follows:
\begin{itemize}
    \item $\forall c \in [\![1, C^+]\!], q \in [\![1,Q]\!]$, $\tilde{p}_s(p_{.,c}^q | \boldsymbol{\dot{X_s}}, \boldsymbol{\dot{C_s}}, \pi) \propto \mbox{Dirichlet}(b_1, ..., b_{K^q})$ with $b_k = \sum_{i = 1}^{n_s}\mathbb{1}_{(\boldsymbol{\dot{C_s}}_i = c)\cap(\boldsymbol{\dot{X_s}}_i^q = k)} + \frac{\beta - 1}{S} + 1$, for all $k$ in $[\![1, K^q]\!]$;
    \item $\tilde{p}_s(\pi | \boldsymbol{\dot{X_s}}, \boldsymbol{\dot{C_s}}, p) \propto \mbox{Dirichlet}(a_1, ..., a_{C^+})$, with $a_c = \sum_{i = 1}^{n_s}\mathbb{1}_{\boldsymbol{\dot{C_s}}_i = c} + \frac{\alpha - 1}{S} + 1$, for all $c$ in $[\![1, C^+]\!]$.
\end{itemize}
where {$\boldsymbol{\dot{C_s}} = (C_{1}, ..., C_{{n_s}})$} are the cluster memberships of the individuals in shard $s$.\\
\newline
This step is embarrassingly parallel over the $S$ shards. In order to accelerate the running time of each MCMC, we implemented an MCMC algorithm in C++17 that is parallelised over the draw of the cluster membership latent variables.\\
\newline
The post-processing steps applied to the MCMC output to identify a single clustering structure are detailed in Appendix \ref{app:MCMC_postprocessing}.

\subsection{The aggregate step in a variational framework}
\label{sec:VCMC_aggregate}

The local posterior distributions are combined to recover the global posterior distribution by assuming: $$ p(\theta|\boldsymbol{\dot{X}}) \approx \prod_{s = 1}^S \tilde{p}_s(\theta|\boldsymbol{\dot{X_s}}) \delta_{\theta = F(\theta^1,...\theta^S)}$$ with F a deterministic aggregation function and $\forall s \in [\![1,S]\!], \theta^s \sim \tilde{p}_s(.|\boldsymbol{\dot{X_s}})$.\\
\newline
Following Rabinovich, Angelino and Jordan (2015) \cite{rabinovich2015}, the choice of the
aggregation function $F$ can be written as a variational inference problem. The optimal choice $F^{opt}$ is chosen to minimise the Kullback-Leibler divergence \cite{KullbackLeibler1951}, $D_{KL}$, between the global posterior distribution $p ( \cdot | \boldsymbol{\dot{X}})$ and its variational approximation $q_F$: 
\begin{equation*}
   F^{opt} = \arg\min_F D_{KL} \big(q_F || p ( \cdot | \boldsymbol{\dot{X}})\big) \mbox{ subject to } q_F \in \mathcal{Q}_F
\end{equation*}
with: $$\mathcal{Q}_{F} = \Big\{ q_F: \theta \mapsto \int_{\Theta^S} \prod_{s=1}^S \tilde{p}_s(\theta^s|\boldsymbol{\dot{X}_s}) \delta_{\theta = F(\theta^1,...,\theta^S)}d\theta^{1:S}, F\in \mathcal{F} \Big\}$$ and $\mathcal{F}$ the set of aggregation functions. \\
\newline
In variational inference, the Kullback-Leibler (KL) divergence between the approximation $q$ and the true posterior can be decomposed as the difference between the log evidence $\log p(\boldsymbol{\dot{X}})$ and the evidence lower bound (ELBO). Since the log evidence does not depend on $q$, minimising the KL divergence is equivalent to maximising the ELBO. In the VCMC setting, however, the ELBO is typically intractable, and \cite{rabinovich2015} therefore propose to maximise a relaxed version $\mathcal{L}$ instead:

\begin{equation}
    F^{opt} = \arg\max_{F} \mathcal{L}(q_{F}) \mbox{ s.t. } q_{F} \in \mathcal{Q}_{F}
\end{equation}
with:
\begin{itemize}
    \item $\mathcal{L}(q_F) = \mathbb{E}_{q_F}[\log p(\theta, \boldsymbol{X})] + \tilde{H}(q_F)$
    \item $\tilde{H}(q_F) = \min_{1 \leq s \leq S} H(\tilde{p}_s) + \frac{1}{S}\sum_{s = 1}^S \mathbb{E}_{\tilde{p}_s}[\log \det J(F^s)(\theta^s)]$ 
    \end{itemize}
Following Rabinovich, Angelino and Jordan (2015) \cite{rabinovich2015}, we restrict the set of aggregation functions to  $\mathcal{F} = \big\{F:  (\theta^1,...,\theta^S) \mapsto \sum_{s = 1}^S \upsilon^s\theta^s\big\}$ with $\forall s, \upsilon^s \in \mathcal{M}_{C \times C}$. This definition of $\mathcal{F}$ fulfils the conditions for which the relaxed variational optimisation problem is convex. \\

\subsection{VCMC for mixture models}
\subsubsection{Matching dependence and the aggregation family}
As mixture models are identifiable only up to a permutation of the cluster labels, we must establish a correspondence between the clusters identified in the different shards. Such a correspondence is called a {\em matching} of the clusters. The set of aggregation functions is actually matching-dependent, and hence we use the notation $\mathcal{F}_a$ for a given matching $a$ in the following. 
More details on clusters matching are given in Section \ref{sec:cluster_alignment}.\\
\newline
We write for all $s$ in $[\![1,S]\!]$, $$\theta^s_{a^s(.)} = (\pi^s_{a^s(.)}, \boldsymbol{p}^s_{a^s(.)}) = \begin{pmatrix} \pi_{a^s(1)}^s & p_{1,{a^s(1)}}^{1,s} & p_{2,{a^s(1)}}^{1,s} & \dots & p_{K^1,{a^s(1)}}^{1,s} & \dots & p_{K^Q,{a^s(1)}}^{Q,s} \\ \vdots & \vdots & \vdots & & \vdots && \vdots \\  \pi_{a^s(c)}^s & p_{1,{a^s(c)}}^{1,s} & p_{2,{a^s(c)}}^{1,s} & \dots & p_{K^1,{a^s(c)}}^{1,s} & \dots & p_{K^Q,{a^s(c)}}^{Q,s} \end{pmatrix} \in \mathcal{M}_{C \times \sum_{q = 1}^Q K^q}$$ the parameters of the local posterior distribution $s$ permuted to achieve matching $a$.
For a given matching $a$, we define $$\mathcal{F}_{a} = \Big\{ F_a: (\theta^1, ..., \theta^S) \mapsto \big(\sum_{s = 1}^S V^s\pi^s_{a(.)}, \sum_{s = 1}^S W^s p_{1, {a(.)}}^{1,s}, \dots, \sum_{s = 1}^S W^s p_{K^1, {a(.)}}^{1,s}, \dots, \sum_{s = 1}^S W^s p_{K^Q, {a(.)}}^{Q,s}\big) \Big\}$$
with:\begin{itemize}
    \item $ \forall s \in \ldbrack 1 , S \rdbrack, \forall q \in \ldbrack 1, Q \rdbrack, \forall k \in \ldbrack 1, K^Q \rdbrack, p_{k, {a(.)}}^{q,s} = \begin{pmatrix}
         p_{k,{a(1)}}^{q,s}\\
         \vdots \\
         p_{k,{a(c)}}^{q,s}\\
    \end{pmatrix}$;
    \item $\forall s \in \ldbrack 1 , S \rdbrack, V^s \in \mathcal{M}_{c \times c}$;\\
    \item $\forall s\in \ldbrack 1 , S \rdbrack, W^s \in \mathcal{M}_{c\times c}$.
\end{itemize}
This definition of $\mathcal{F}_a$ fulfils the conditions for which the relaxed variational optimisation problem is convex \cite{rabinovich2015}.\\
\newline

\subsubsection{Constraint-preserving aggregation weights}
In order to preserve the constraints of the categorical mixture model, namely that the mixture weights and the parameters of the cluster-specific distributions sum to 1, some constraints on $(V^s, W^s)_{s \in [\![1,S]\!]}$ must be imposed. \\
\newline
Let us define $\hat{\theta} = F_a(\theta^1, ..., \theta^S)$. In order to meet the following constraints: \begin{enumerate}
    \item $\sum_{c = 1}^C \hat{\pi}_c = 1 $;
    \item $\forall q \in \ldbrack 1, Q \rdbrack , \forall c \in \ldbrack 1, C \rdbrack, \sum_{k = 1}^{K^q} \hat{p}_{k,c}^q = 1$;
\end{enumerate}
we set:\\
\begin{enumerate}
\item $\forall s \in \ldbrack 1, S \rdbrack, V_s = \lambda^s I_C$ where $I_C$ stands for the identity matrix of size $C \times C$ with $\lambda^s \in [0,1]$ and $\sum_{s = 1}^S \lambda^s = 1$. \\
\item  $\forall s \in \ldbrack 1, S \rdbrack, W^s = \begin{pmatrix}
    \mu_1^s & & \\
    &\ddots &  & \mbox{\Huge 0} &\\
    &\mbox{\Huge 0}  & &\ddots & \\
    && &&\mu_C^s
\end{pmatrix}$.\\
\end{enumerate}
In the following, we will refer to $(\lambda^s, \mu_1^s, ... \mu_C^s)_{s\in [\![1,S]\!]}$ as aggregation weights.\\
\newline
One can note that the aggregation weights on $\pi$, $(\lambda^s)_{s\in [\![1,S]\!]}$, depend only on the shards, meaning that the contribution of a given shard on the global estimates of the cluster weights is the same for all the cluster weights. The aggregation weights on $\boldsymbol{p}$, $(\mu_1^s, ... \mu_C^s)_{s\in [\![1,S]\!]}$, depend on the clusters and the shards. If the distribution of the clusters in a given shard is not representative of the distribution of the clusters across the full data, the contribution of this shard to the global estimates of the parameters of the cluster-specific distributions can be higher for the clusters that are over-represented and smaller for the clusters that are under-represented.\\
\newline
\subsubsection{Optimisation via projected stochastic gradient descent}
The variational problem is now simplified to the identification of aggregation weights that minimise the relaxed evidence lower bound. This is done using a projected stochastic gradient descent algorithm. This algorithm is performed in two steps: \begin{enumerate}[label=(\alph*)]
    \item a stochastic gradient descent in the unconstrained space $\mathcal{M}_{C \times \sum_{q = 1}^Q K^q}(\mathbb{R})$;
    \item the projection of the output of the first step in the constrained space of the aggregation weights with Algorithm 1 from Duchi \textit{et al.} (2008) \cite{Duchi2008}.
\end{enumerate} 
We detail four strategies to perform the stochastic gradient descent, each appropriate for different Federated Learning settings. The first three ones can be applied to infer any statistical model with VCMC, but the last one is specific to mixtures of categorical distributions.
\begin{enumerate}
    \item The first option, used by Rabinovich, Angelino and Jordan (2015) \cite{rabinovich2015}, consists of sharing the data to the global node before performing the projected stochastic gradient descent. This option is the least computationally demanding and should be preferred when VCMC is only used to accelerate the MCMC inference, but not to prevent data sharing.
    \item The gradient descent can be performed in a Federated Learning way, by computing the gradient descent direction locally and computing the global direction by summing the local directions at the global level. This option should be used if the full data cannot be stored on a single node (because of size or privacy issues), but unlimited communication between the local nodes and the global node is available. 
    \item If the full data cannot be stored on the global node and the communication between the nodes is limited to a maximum number $M$  (for example, if the communication has to go through human operators), $L$ federated learning stochastic gradient descents with $M$ steps can be performed in parallel. Only the output of the run leading to the greatest ELBO is kept. The smaller $M$ is the larger $L$ should be. In order to maximise the chance of finding the global optimum, the space of starting points should be covered approximately uniformly by the $L$ starting points. To do so, one can uniformly sample $L$ points, $\{U_1, ..., U_L\}$, in the simplex of size $(S-1) \times (C+1)$ following a Latin Hypercube Sampling design \cite{McKay1979}. Then, use the Rosenblatt transformation \cite{Rosenblatt1952, ZhangZhangZhou2020} to project the $L$ points into the space $\mathcal{X'} = \{ M \in \mathcal{M}_{(S - 1)\times (C+1)}, \forall i \in [\![1,S]\!], \forall j \in [\![1,C+1]\!], m_{ij} \in [0,1] \mbox{ and } \forall j \in [\![1, C+1]\!], \sum_{i = 1}^{S} m_{ij} \leq 1\}$. We will write $\{M'_1, ..., M'_L\}$ those new points. In our case, the Rosenblatt transformation is available in closed form: 
        $$m'_{ij} = \left\{\begin{array}{ll} (1 - (1 - u_{i + (S-1) \times (j - 1)})^{\frac{1}{S-1}}) & \mbox{ if } i = 1\\ (1 - \sum_{i' = 1}^{i - 1} m'_{i'j})[1 - (1 - u_{i + (S-1) \times (j - 1)})^{\frac{1}{S- i}}] & \mbox{ else } .
\end{array}
\right.$$\\
For each point $l$, project $M'_l$ into the space $\mathcal{X} = \{ M \in \mathcal{M}_{(S - 1)\times (C+1)}, \forall i \in [\![1,S]\!], \forall j \in [\![1,C+1]\!], m_{ij} \in [0,1], \forall j \in [\![1, C+1]\!], \sum_{i = 1}^{S} m_{ij} = 1\}$. Such a point is referred to as $M_l$ and is defined with the following equation: $$\forall j \in [\![1, C+1]\!], m_{ij} = \left\{
\begin{array}{ll}
m'_{ij} & \mbox{ if } i < S \\
1 - \sum_{i = 1}^{S-1} m'_{ij} & \mbox{ if } i = S .
\end{array}
\right.$$
    \item The last option is to share to the global node a summary of the data shards. This summary, that will be referred to as contingency matrix for brevity, is the number of individuals presenting each modality for each covariate, $\{\#(X^1 = k_1 \cap X^2 = k_2 \cap \cdots \cap X^Q = K^q), K^q \in [\![1,K^Q]\!], q \in [\![1,Q]\!]\} $ where the notation $\#A$ stands for the cardinality of the set $A$. The equation involving the data in the gradient descent can be written as follows:
    \begin{align*}
    \log p(\theta,\boldsymbol{X}) 
    &= \sum_{k_1 = 1}^{K^1}\cdots \sum_{k^Q = 1}^{K^Q} \#(X^1 = k_1 \cap X^2 = k_2 \cap \cdots \cap X^Q = K^q)\log \sum_{c = 1}^C \pi_c \prod_{q = 1}^Q p_{k^q, c}^q \\ 
    &+(\alpha - 1) \sum_{c = 1}^C \log \pi_c + (\beta - 1) \sum_{q = 1}^Q \sum_{c = 1}^C \sum_{k = 1}^{K^q} \log p_{k,c}^q - \log B(\alpha) - (Q +C)\log B(\beta).
\end{align*} The decision to share the contingency matrix must be taken after consulting the data as it does not necessarily preserve privacy. Indeed, for large $Q$ it is not uncommon that most of the individuals present a unique combination of covariates' modalities. 
\end{enumerate}

\subsubsection{Cluster matching}
\label{sec:cluster_alignment} 
The mixture model defined in Section \ref{sec:StatModel} is invariant by a permutation of the cluster labels. This means that the cluster labels might not correspond from one shard to another. Before being able to perform the aggregate step, one must ensure the correspondence of the cluster labels in the different shards. This step is referred to as matching of the clusters. To do this, we consider three different options :
\begin{enumerate}
    \item {\bf Hungarian matching:} using the Hungarian algorithm as presented by Rabinovich, Angelino and Jordan (2015) \cite{rabinovich2015};
    \item {\bf Minimum divergence matching:} choosing the matching that minimises the Kullback-Leibler divergence between the local marginal posterior distributions of the parameters of the cluster-specific distributions;
    \item {\bf Ball matching:} grouping the clusters that are ``close" to each other.
    \end{enumerate}
A detailed description of the three methods is provided below. All the methods are described for any distance $d_l$ on $\mathbb{R}^l, l \in \mathbb{N}$. In practice, we set $d_l$ to the Euclidean distance on $\mathbb{R}^l$. 

\paragraph{Hungarian matching}
The method is described below for $S = 2$, but it can be easily generalised to $S>2$ by iteratively repeating the matching to a reference partition. A good choice for this reference partition, is any partition of maximum size derived from the local posterior distributions. \\
\newline
We consider a distance $d_{\sum_{q =1}^QK^q}$ between two elements of $\mathbb{R}^{\sum_{q =1}^QK^q}$. Let $\hat{C}^s$ denote the estimated number of clusters in shard $s$, $s \in \{1,2\}$.\\
The following alignment problem is solved using the Hungarian algorithm: $$(y_{c_1,c_2})_{(c_1, c_2) \in [\![1,\hat{C}_1]\!]\times [\![1,\hat{C}_2]\!]} = \arg\min \sum_{c_1 = 1}^{\hat{C}_1} \sum_{c_2 = 1}^{\hat{C}_2} y_{c_1,c_2}d(\boldsymbol{p}^1_{c_1}, \boldsymbol{p}^2_{c_2})$$
$$\mbox{ s.t } \forall  (c_1, c_2) \in [\![1,\hat{C}_1]\!] \times [\![1,\hat{C}_2]\!], y_{c_1,c_2} \in \{0,1\}, \sum_{c_1 = 1}^{\hat{C}_1} y_{c_1,c_2} = 1, \sum_{c_2 = 1}^{\hat{C}_2} y_{c_1,c_2} = 1 .$$
The identified set defines the match of the clusters in shard 1 and shard 2.

\paragraph{Minimum divergence matching}
We build on the work of Stephens (2000) \cite{Stephens2000} for matching cluster labels within different MCMC chains.\\
\newline
Let us define $\boldsymbol{P}(\boldsymbol{\dot{X}}_s;\boldsymbol{p}_{a(.)}^s) \in \mathcal{M}_{n_s \times C}$ with $C \in \mathbb{N}^*$ such that  
$$\forall i \in[\![1,n_s]\!], \forall j \in [\![1,C]\!], \big(\boldsymbol{P}(\boldsymbol{\dot{X}}_s;\boldsymbol{p}_{a(.)}^s)\big)_{ij} = \mathbb{P}(C_i = j|X_i, \boldsymbol{p}_{a(.)}^s)$$
It is straightforward to show that $\forall i \in[\![1,n_s]\!], \forall j \in [\![1,C]\!], \mathbb{P}(C_i = j|X_i, \boldsymbol{p}_{a(.)}^s) = \frac{\mathbb{E}_{\pi}[\pi_j] \mathbb{P}(X_i|p_j)}{\sum_{k = 1}^C \mathbb{E}_{\pi}[\pi_k] \mathbb{P}(X_i|p_k)}$.\\
\newline
We start by describing the method for $S = 2$.\\ Two partitions are matched by choosing $\hat{a}(.)$ that solves $$\hat{a}(.) = \arg\min_{a(.)}\mathbb{E}_{p(\boldsymbol{p}^1,\boldsymbol{p}^2|\boldsymbol{\dot{X}})}\Big[D_{KL}\Big(\boldsymbol{P}(\boldsymbol{\dot{X}}_2;\boldsymbol{p}_{a(.)}^2)||\boldsymbol{P}(\boldsymbol{\dot{X}}_2;\boldsymbol{p}_{\epsilon(.)}^1)\Big)\Big]$$ where $\epsilon(.)$ is the identity permutation.\\
\newline
In practice, solving this problem is not straightforward as there exists $\sum_{c_1 = 1}^{C^1}\sum_{c_2 =1}^{min(c_1, C^2)} \binom{C^1}{c_1}\binom{C^2}{c_2}$ different combinations, with $C^s$ the number of clusters in shard $s$, $s \in \{1,2\}$. We propose a workflow to simplify the problem to an alignment problem. In the two shards, we consider a mixture model of size $C^m = \max(C^1, C^2) + 1$. The parameters of the distribution of the extra empty clusters are assumed to follow the prior distribution. In this setting, the problem can be written as an alignment problem where the final cost is equal to $\mathbb{E}_{p(\boldsymbol{p}^1,\boldsymbol{p}^2|\boldsymbol{\dot{X}})}\Big[D_{KL}\Big(\boldsymbol{P}(\boldsymbol{\dot{X}}_2;\boldsymbol{p}_{\hat{a}(.) }^2)||\boldsymbol{P}(\boldsymbol{\dot{X}}_2;\boldsymbol{p}_{\epsilon(.)}^1)\Big)\Big] = \sum_{c = 1}^{C^m}\mathbb{E}_{p(\boldsymbol{p}^1,\boldsymbol{p}^2|\boldsymbol{\dot{X}})}\Big[D_{KL}\Big(\boldsymbol{P}(\boldsymbol{\dot{X}}_2;\boldsymbol{p}_{\hat{a}(c) }^2)||\boldsymbol{P}(\boldsymbol{\dot{X}}_2;\boldsymbol{p}_{\epsilon(c)}^1)\Big)\Big]$. As there is at least one empty cluster in each shard, there exists $c \in [\![1 ,C^m]\!]$, such that $\boldsymbol{p}_{\hat{a}(c)}^2$ and $\boldsymbol{p}_{c}^1$ are drawn from the prior distribution. In this case, $\mathbb{E}_{p(\boldsymbol{p}^1,\boldsymbol{p}^2|\boldsymbol{X})}\Big[D_{KL}\Big(\boldsymbol{P}(\boldsymbol{X}_2;\boldsymbol{p}_{\hat{a}(c) }^2)||\boldsymbol{P}(\boldsymbol{X}_2;\boldsymbol{p}_{\epsilon(c)}^1)\Big)\Big]$ can be interpreted as the cost of matching two clusters at random. We decide to "unmatch" all pairs of clusters indexed by $(c', a(c')), c' \in [\![1,C^m]\!]\setminus\{c\}$ such that $$\mathbb{E}_{p(\boldsymbol{p}^1,\boldsymbol{p}^2|\boldsymbol{X})}\Big[D_{KL}\Big(\boldsymbol{P}(\boldsymbol{X}_2;\boldsymbol{p}_{\hat{a}(c') }^2)||\boldsymbol{P}(\boldsymbol{X}_2;\boldsymbol{p}_{\epsilon(c')}^1)\Big)\Big] \geq \mathbb{E}_{p(\boldsymbol{p}^1,\boldsymbol{p}^2|\boldsymbol{X})}\Big[D_{KL}\Big(\boldsymbol{P}(\boldsymbol{X}_2;\boldsymbol{p}_{\hat{a}(c) }^2)||\boldsymbol{P}(\boldsymbol{X}_2;\boldsymbol{p}_{\epsilon(c)}^1)\Big)\Big].$$\\
\newline
In order to align the clusters from $S$ shards, we repeat iteratively the process. Details can be found in Algorithm \ref{alg:KLmatching}.

\begin{algorithm}
    \caption{Minimum divergence matching}
    \label{alg:KLmatching}
    \begin{algorithmic}
        \State \textbf{Output: } $a_1(.),...,a_S(.)$
        \State $\boldsymbol{p}^* = \boldsymbol{p}^1$
        \State $w_1 = (1, ..., 1) \in \mathbb{R}^{C^1}$
        \State $C^* = C^1 $
        \State $a_1(.) = \epsilon(.)$
        \For{$s \in [\![2,S]\!]$}
        \State Set $C^m = \max(C^*, C^s) + 1$
        \State Find $\hat{a}_s(.) = \arg \min_{a(.)}\mathbb{E}_{p(\boldsymbol{p}^*,\boldsymbol{p}^s|\boldsymbol{X})}\Big[D_{KL}\Big(\boldsymbol{P}(\boldsymbol{X}_s;\boldsymbol{p}_{a(.)}^s)||\boldsymbol{P}(\boldsymbol{X}_s;\boldsymbol{p}_{\epsilon(.)}^*)\Big)\Big]$
        \State Identify $c \in [\![1 ,C^m]\!]$, such that $\boldsymbol{p}_{\hat{a}_s(c)}^s$ and $\boldsymbol{p}_{c}^*$ are drawn from the prior distribution.
        \For{ each element in $\{c' \in [\![1,C^m]\!]\setminus\{c\}, \mathbb{E}_{p(\boldsymbol{p}^*,\boldsymbol{p}^s|\boldsymbol{X})}\Big[D_{KL}\Big(\boldsymbol{P}(\boldsymbol{X}_s;\boldsymbol{p}_{\hat{a}_s(c') }^s)||\boldsymbol{P}(\boldsymbol{X}_s;\boldsymbol{p}_{\epsilon(c')}^*)\Big)\Big] \geq \mathbb{E}_{p(\boldsymbol{p}^*,\boldsymbol{p}^s|\boldsymbol{X})}\Big[D_{KL}\Big(\boldsymbol{P}(\boldsymbol{X}_s;\boldsymbol{p}_{\hat{a}_s(c) }^s)||\boldsymbol{P}(\boldsymbol{X}_s;\boldsymbol{p}_{\epsilon(c)}^*)\Big)\Big]\}$}
        \State Update $a_s(.)$ such that $a_s(c') = C^m$.
        \State $C^m = C^m + 1$
        \EndFor 
        \State Compute $w_s \in \mathbb{R}^{C^m - 1}$ where  $(w_{s,c})^{-1} =  \left\{
\begin{array}{ll}(w_{s-1,c})^{-1} + 1 & \mbox{if } a_s(c) \in [\![1,C^*]\!] \;\\
1 & \mbox{otherwise .}
\end{array}\right. $
        \State Update $\boldsymbol{p}^* = \frac{w_{s}}{w_{s-1}}\boldsymbol{p}^* + w_s\boldsymbol{p}^s_{a_s(.)}$
        \State Update $C^* = C^m - 1$
        \EndFor
    \end{algorithmic}
\end{algorithm}

\paragraph{Ball matching}
\begin{figure}
    \centering
    \includegraphics[width=1\linewidth]{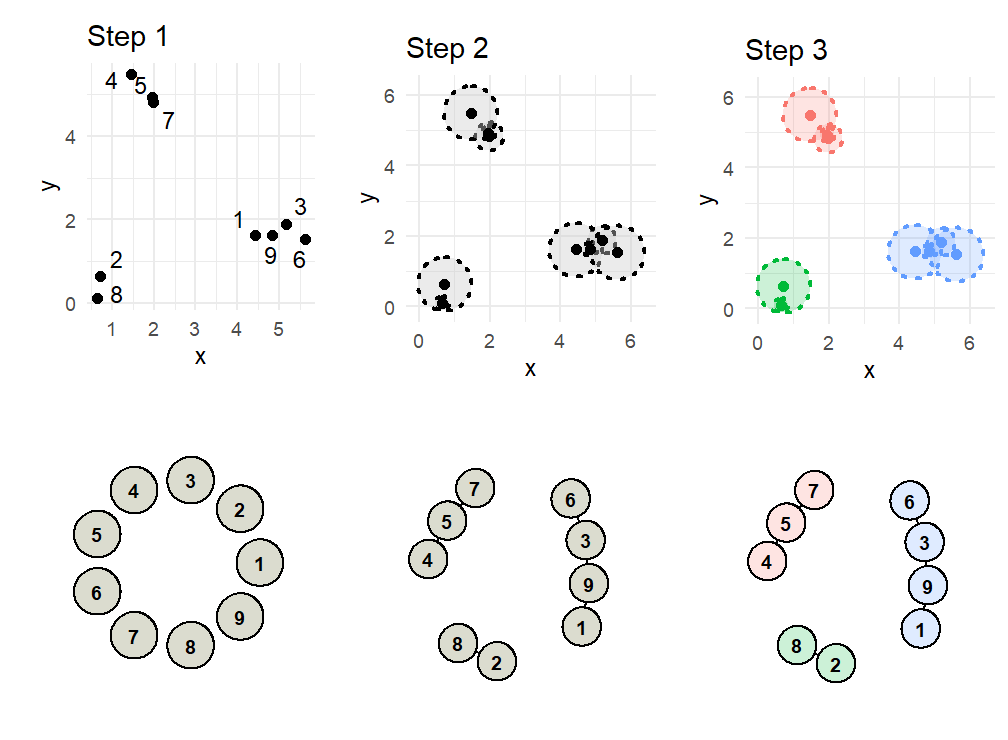}
    \caption{Diagram of the Ball matching procedure : step 1 is the identification of the clusters coordinates, step 2 the identification of the close clusters, step 3 the identification of the groups of close clusters.}
    \label{fig:Diag_ball_matching}
\end{figure}
In the spirit of the work of Guha, Ho and Nguyen (2021) \cite{GuhaHoNguyen2021} for the inference of Dirichlet process mixture models, the idea is to match clusters that are ``close" to each other. Cluster $a$ is said to be $r_a$-close to cluster $b$, if the parameters of the cluster-specific distributions of cluster $b$ lie in a ball of radius $r_a$ centred on the parameters of the cluster-specific distributions of cluster $a$. A cluster $a$ is said to be {\em connected} to cluster $b$ if cluster $b$ is in the set of $r_a$-close clusters. Iteratively for each cluster, we identify the set of connected clusters. If no $r_a$-close cluster exists for the considered cluster, then no connected cluster is defined. Then, we can define a simple undirected graph where the nodes are the different clusters and there exists an edge between nodes $a$ and $b$ if and only if the cluster represented in node $a$ (resp. $b$) is connected to the cluster represented in node $b$ (resp. $a$). All the clusters within each connected component of the graph are matched together. Figure \ref{fig:Diag_ball_matching} provides a schematic description of the procedure. More details on the algorithm are provided in Algorithm \ref{alg:ball_matching}.\\
\newline
The output of the algorithm depends on the choice of $r$. According to Nguyen (2013) \cite{Nguyen2013} and Ho and Nguyen (2016)\cite{HoNguyen2016}, in the setting of over-fitted mixture models (under specific conditions), posterior distribution of the cluster-specific distribution parameters are converging to the truth in Euclidean norm with maximum rate $(\frac{\log n}{n})^{\frac{1}{4}}$ where $n$ is the number of observations. Consequently, for the cluster $c$ in shard $s$, we set $r_c$ to be equal to  $(\frac{\log n_s}{n_s})^{\frac{1}{4}}$. \\
\newline
\begin{algorithm}
    \caption{Ball matching}
    \label{alg:ball_matching}
    \begin{algorithmic}
        \State \textbf{Output:} The connected components of $\mathcal{G}$
        \State $\mathcal{N} = \{\boldsymbol{p}_{1}^1, ..., \boldsymbol{p}_{C_1}^1,..., \boldsymbol{p}_{1}^S, ..., \boldsymbol{p}_{C_S}^S \}$
        \State $\mathcal{E} = \{\}$
        \State $ \omega_{n_{s(c)}} = (\frac{\log n_{s(c)}}{n_{s(c)}})^{\frac{1}{4}}$ where $s(c)$ is the shard in which cluster $c$ was inferred.
        \For{$\boldsymbol{p}_c \in \mathcal{N}$}
        \State Compute $\mathcal{C} = \{ {j \in \mathcal{N}\setminus \{c\}}:\lVert \boldsymbol{p}_c - \boldsymbol{p}_j \rVert < \omega_{n_{s(c)}} \}$ 
        \State $\mathcal{E} \leftarrow \mathcal{E} \cup \{(c, j) : j \in \mathcal{C}\}$
        \EndFor   
        \State Find the connected components of $\mathcal{G} = \{\mathcal{N}, \mathcal{E}\}$
    \end{algorithmic}
\end{algorithm}
\noindent
In the case of mixtures of categorical distributions, the parameters of the cluster-specific distribution on the $q$\textsuperscript{th} variable lies in the probability simplex of size $K^q$, which is a space of dimension $K^q - 1$. So for each $q$, we consider the $K^q - 1$ first components of the parameters of the cluster-specific distribution and the considered distance is a distance in  $\mathbb{R}^{\sum_{q = 1}^Q(K^q - 1)}$.\\
\newline
Code implementing the VCMC framework to infer over-fitted Bayesian mixture models of categorical distributions is available at \textcolor{red}{(public GitHub in progress)}.

\section{Simulation study}
\label{sec:simulation_study}
We perform a simulation study in order to compare the impact of the different algorithmic options and to assess the performance of VCMC for over-fitted mixture models.

\subsection{Simulation design}
For each simulation scenario, we generate 100 datasets composed of 4000 individuals that are split into 4 shards (each shard containing 1000 individuals). We consider three different options to split the data into shards.\\
The first option consists of a homogeneous distribution of the clusters across the shards. Data are generated with five underlying clusters. These five clusters are uniformly distributed across the four shards. \\
We refer to the second considered option as heterogeneous nested distribution. In this setting, data are generated with six underlying clusters. The observations from the six clusters are observed all together only in the first shard. The second shard contains observations belonging to five of the six clusters. The third shard contains observations belonging to three of the six clusters, and the fourth shard contains observations belonging only to two of the six clusters. The distribution of the clusters is uniform within each shard. However, this distribution is not uniform across the shards.\\
The last considered setting is the heterogeneous non-nested distribution. Data are generated with six underlying clusters. None of the shards contains observations from all of the six clusters. Clusters 1 to 5 are observed in the first shard. The second shard contains observations belonging to clusters 2 to 6. It is the only shard that contains observations from cluster 6. The third shard contains observations belonging to clusters 1 to 3, and the fourth shard only contains observations belonging to clusters 1 and 2. The distribution of the clusters is uniform within each shard. However, this distribution is not uniform across the shards.\\ Table \ref{tab:proba_simulated_clusters} summarises the probability distributions of the clusters within the shards and in the full data in the three settings described above.\\

\begin{figure}[htbp]
     \centering
     \begin{subfigure}[b]{0.45\textwidth}
         \centering
         \includegraphics[width=\textwidth]{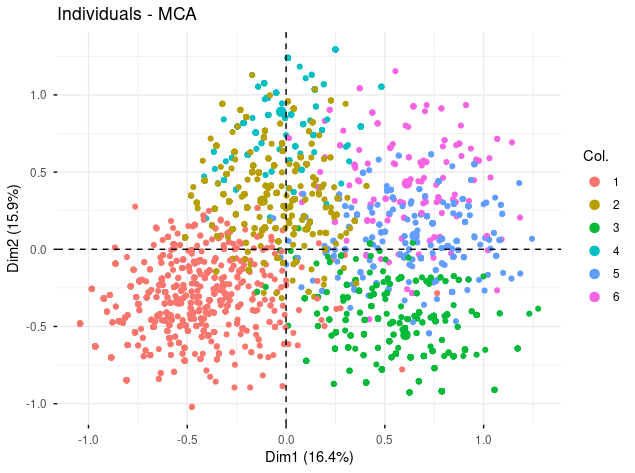}
         \caption{Easy separability}
         \label{fig:MCAeasy}
     \end{subfigure}
     \hfill
     \begin{subfigure}[b]{0.45\textwidth}
         \centering
         \includegraphics[width=\textwidth]{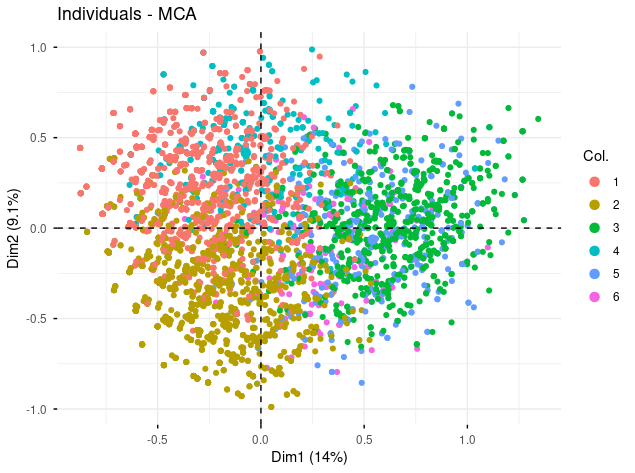}
         \caption{Poor separability}
         \label{fig:MCApoor}
     \end{subfigure}
     \caption{MCA coordinates on the first and second axes of the observations in a simulated dataset in the setting of a heterogeneous nested distribution of the clusters}
     \label{fig:MCAboth}
\end{figure}

\begin{table}[H]
    \centering
    \begin{tabular}{|c|c||c|c|c|c|c|c|}
    \cline{2-8}
        \multicolumn{1}{c|}{}& \backslashbox{Shard}{Cluster} &  1 & 2 & 3 & 4 & 5 & 6\\
        \cline{2-8}
        \hline
        \multirow{5}{3cm}{homogeneous shards} & 1 & 1/5 & 1/5 & 1/5 & 1/5  & 1/5 & 0 \\
        & 2 & 1/5 & 1/5 & 1/5 & 1/5  & 1/5 & 0\\
        & 3 & 1/5 & 1/5 & 1/5 & 1/5  & 1/5 & 0\\
        & 4 & 1/5 & 1/5 & 1/5 & 1/5  & 1/5 & 0 \\
        \cline{2-8}
        & Global& 0.20 & 0.20 & 0.20 & 0.20  & 0.20 & 0 \\
        \hline
        \multirow{5}{3cm}{heterogeneous nested shards} & 1 & 1/6 & 1/6 & 1/6 & 1/6  & 1/6 & 1/6 \\
        & 2 & 1/5 & 1/5 & 1/5 & 1/5  & 1/5 & 0\\
        & 3 & 1/3 & 1/3 & 1/3 & 0  & 0 & 0\\
        & 4 & 1/2 & 1/2 & 0 & 0 & 0 & 0\\
        \cline{2-8}
        & Global& 0.30 & 0.30 & $\approx$ 0.18 & $\approx$ 0.09 & $\approx$ 0.09 & $\approx$ 0.04\\
        \hline
        \multirow{5}{3cm}{heterogeneous non-nested shards} &1 & 1/5 & 1/5 & 1/5 & 1/5 & 1/5 & 0 \\
        & 2 & 0 & 1/5 & 1/5 & 1/5  & 1/5 & 1/5\\
        & 3 & 1/3 & 1/3 & 1/3 & 0  & 0 & 0\\
        & 4 & 1/2 & 1/2 & 0 & 0 & 0 & 0\\
        \cline{2-8}
        &Global& $\approx$ 0.26 & $\approx$ 0.31 & $\approx$ 0.18 &  0.10 &  0.10 & 0.05\\
        \hline    
    \end{tabular}
    \caption{Distribution of the clusters across the shards and in the full data (global) for the three simulation settings}
    \label{tab:proba_simulated_clusters}
\end{table}
\noindent
For each of the three settings described previously, the observations are characterised by ten categorical covariates (four with two modalities and six with three modalities) having cluster-specific distributions. The distribution of the parameters of the cluster-specific distributions influences the separability of the clusters. For the purpose of our study, we define two settings. In the first setting, that leads to easy separability of the clusters based on the covariates, we draw the parameters of the cluster-specific distribution from a symetric Dirichlet distribution $\mathcal{D}(0.3)$. The second setting leads to a poor separability of the clusters based on the covariates. For each cluster and each covariate, the parameters of cluster-specific distributions are drawn from a symetric Dirichlet distribution $\mathcal{D}(1)$. Figure \ref{fig:MCAboth} shows a two-dimensional representation of a dataset with the two previously described settings for cluster separability in the case of a heterogeneous nested distribution of the clusters across the shards.

\noindent
For all the simulation scenarios, we run our algorithm with the following settings. We set $C^+$ to 15 in the statistical model. During the apply phase the MCMC algorithm is run for 50,000 iterations including 25,000 of burn-in. The initialisation of the MCMC algorithm (parameters and latent variables) is at random. Unless stated otherwise, we perform the aggregate phase by sharing the contingency matrix with the global node. We perform the stochastic gradient descent until the difference in ELBO between two iterations becomes smaller than 0.1 with a maximum of 1000 iterations.

\subsection{Matching algorithm comparison}
\label{sec:matching_comparison}

In order to compare the matching algorithms, we study two questions: how well does the algorithm recover the clustering structure in the data; how close are the point estimates of the model parameters derived from the global posterior density to the true model parameters.
The answer to the first question is assessed by comparing the estimated number of clusters to the true number of clusters and by studying the adjusted Rand index \cite{HubertArabie1985}. The answer to the second question is studied through the empirical relative bias and the 95\% coverage rate of the point estimates of the model parameters.\\
\newline
\begin{table}[]
    \centering
    \resizebox{\textwidth}{!}{
    \begin{tabular}{|c|c|c|c|c|}
    \hline
        Type of shards & Cluster separability & Matching method & Correct matching recovered & mean ARI \\
        \hline
         \multirow{6}{*}{Homogeneous}& \multirow{3}{*}{easy} & Hungarian & 61\% & 0.88\\
         \cline{3-5}
         & & Minimum divergence & 61\% & 0.96 \\
         \cline{3-5}
         & & Ball & 61\% & 0.96 \\
         \cline{2-5}
         & \multirow{3}{*}{poor}  & Hungarian & 69\% & 0.57 \\
         \cline{3-5}
         & & Minimum divergence & 69\%& 0.62 \\
         \cline{3-5}
         & & Ball & 71\%\ & 0.62\\
         \hline
         \multirow{6}{*}{heterogeneous nested}& \multirow{3}{*}{easy} & Hungarian & 63\% & 0.97\\
         \cline{3-5}
         & & minimum divergence & 55\% & 0.96 \\
         \cline{3-5}
         & & Ball & 58\% & 0.97 \\
         \cline{2-5}
         & \multirow{3}{*}{poor}  & Hungarian & 71\% & 0.73\\
         \cline{3-5}
         & & minimum divergence & 71\% & 0.72\\
         \cline{3-5}
         & & Ball & 73\% & 0.73 \\
         \hline
         \multirow{6}{*}{heterogeneous non-nested}& \multirow{3}{*}{easy} & Hungarian & 0\% & 0.90\\
         \cline{3-5}
         & & minimum divergence & 58\% & 0.97 \\
         \cline{3-5}
         & & Ball & 58\% & 0.98 \\
         \cline{2-5}
         & \multirow{3}{*}{poor}  & Hungarian & 0\% & 0.69\\
         \cline{3-5}
         & & Minimum divergence & 74\% & 0.76\\
         \cline{3-5}
         & & Ball & 76\% & 0.77\\
         \hline
    \end{tabular}
    }
    \caption{Percentage of simulated data for which the correct matching is recovered and mean adjusted Rand index (ARI) between the estimated partition and the true partition on the full data when the partition in the local shards is estimated for 100 simulated datasets in different data simulation scenarios.}
    \label{tab:matching_comparison_estimated_C}
\end{table}
\begin{figure}
    \centering
    \includegraphics[width=1\linewidth]{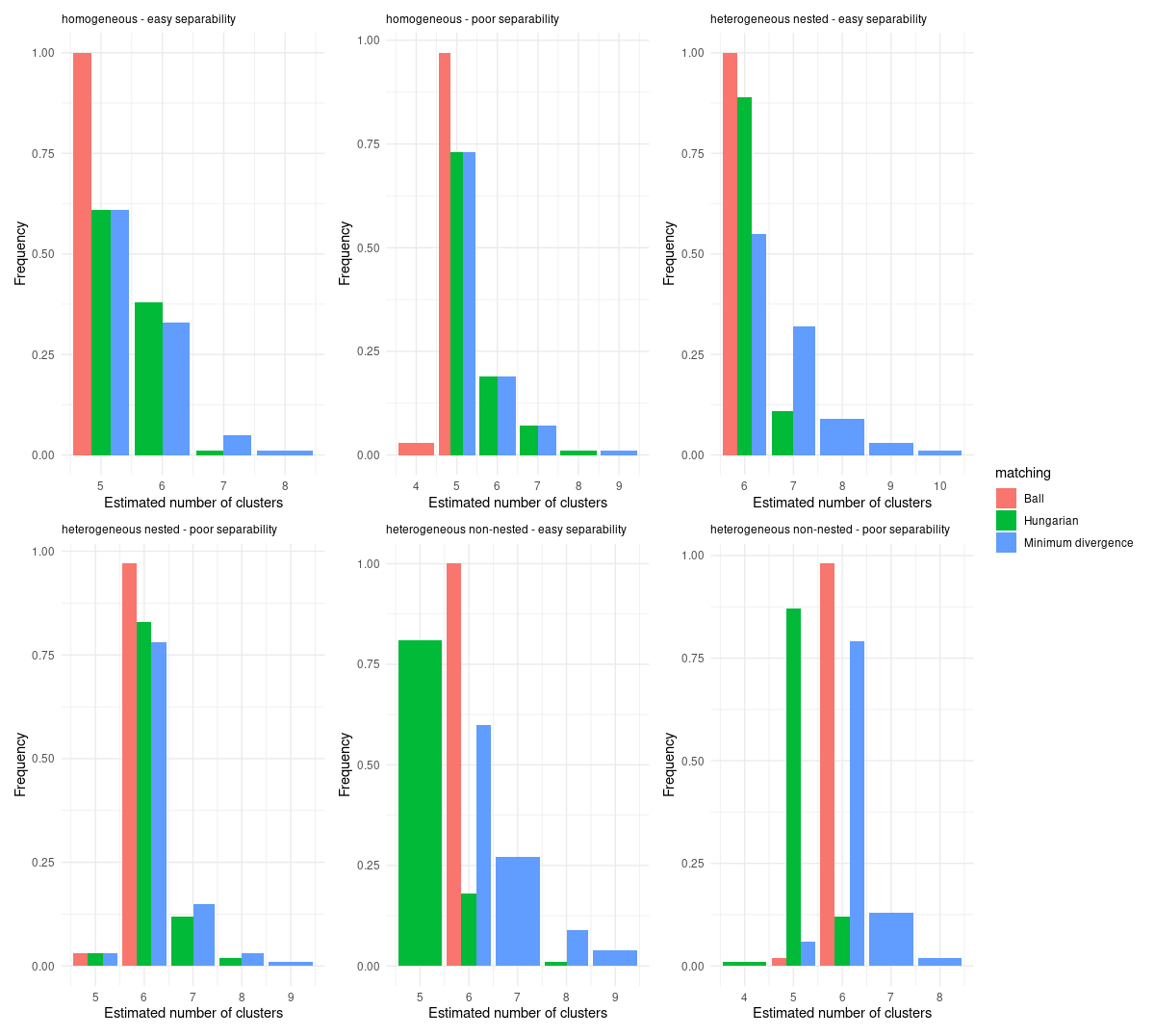}
    \caption{Distribution of the estimated number of clusters in the full data for the three matching algorithms estimated with 100 simulated datasets in different data simulation scenarios.}
\label{fig:barplot_n_clust_matching_comparison_estimated_C}
\end{figure}
\noindent
We compare the three methods described in Section \ref{sec:cluster_alignment} to perform the cluster matching. Table \ref{tab:matching_comparison_estimated_C} shows the percentage of simulated data for which the correct matching is recovered and mean adjusted Rand index (ARI) between the estimated partition and the true partition on the full data when the partition in the local shards is estimated for 100 simulated datasets in different data simulation scenarios. In the case of a homogeneous distribution of the easily separable clusters across the shards, the Hungarian, the minimum divergence and the Ball matchings recover the true partition in the same proportions (61\%). In the case of poor cluster separability, the performances of the Hungarian and the minimum divergence matchings are similar. The Ball matching slightly outperforms the two other methods, i.e. the correct matching is recovered in 71\% of the simulated datasets by the former and 69\% by the latter. In the case of a heterogeneous nested clusters distribution across the shards, the Hungarian matching recovers more often the correct matching than the minimum divergence matching or the Ball matching for easily separable cluster (respectively for 63\%, 55\%  and 58\% of the simulated datasets). However, the Hungarian and the minimum divergence matching algorithms perform equally in terms of recovering the true matching for poorly cluster separability and are slightly outperformed by the Ball matching algorithm (the true matching is recovered in 71\% of datasets for the former and 73\% for the latter). For scenarios with non-nested distribution of the clusters across the shards, the Hungarian matching never recovers the correct matching. This result is expected because the Hungarian algorithm enforces a one to one correspondence of all the clusters across the different shards when the number of clusters is the same. Surprisingly, the performances of the minimum divergence matching and the Ball matching are slightly better in this case than in the case of an nested heterogeneous repartition of the clusters across the shards. The two latter algorithms perform identically in the case of easily separable clusters (the true matching is recovered for 58\% of the simulations), whereas the Ball matching outperforms slightly the minimum divergence matching for poorly separable clusters (the true matching is respectively recovered for 74\% and 76\% of the simulated datasets).
This simulation study does not allow to provide clear guidelines on whether to use the minimum divergence matching or the Ball matching. However it emphasises that the Hungarian matching should not be used without the guarantee that all the clusters are observed in at least one of the shards. Furthermore, this shard should be clearly identified for its estimated partition to be used as the reference for the matching algorithm. \\
\newline
Figure \ref{fig:barplot_n_clust_matching_comparison_estimated_C} shows the distribution of the estimated number of clusters in the full data for the three matching algorithm estimated with 100 simulated datasets in different data simulation scenarios. As the Hungarian matching has to match as many clusters as possible, its results provide information on the maximum number of clusters estimated locally. It seems that the Ball matching tends to match clusters intra-shard when needed, whereas the minimum divergence matching overestimates quite often the number of clusters. \\
Obviously, the recovery of the correct matching depends on the quality of the estimated local partitions. This remark motivates a second study for which the estimated partitions in each shard are set to the truth. Results are presented in Appendix \ref{app:cluster_matching}. \\
\newline
Because the Ball matching seems to be able to mitigate the error on the estimated number of clusters produced by local MCMC algorithm, this matching option will be used in the remainder of this work.\\

\subsection{Inference strategy comparison}
We compare two inference strategies: both use a stochastic gradient descent (any derivation of it as the stochastic federated gradient descent or the stochastic gradient descent on the contingency matrix will lead to the same results as no approximation is required) but the first one considers 3 random starting points and a maximum of 1000 iterations whereas the second one is set with 300 starting points drawn following the third inference procedure described in Section \ref{sec:VCMC_aggregate} and a maximum of 10 iterations. The maximum number of iterations over all the starting points is the same in the two settings. For the two inference strategies, we only keep the run leading to the maximum ELBO. \\
\newline
When applying to 100 datasets simulated with a heterogeneous distribution of the clusters across the shards and easy cluster separability, we find that the algorithm with 300 starting points and a maximum of 10 iterations achieves a higher ELBO in 97\% of the cases compared to the algorithm with 3 starting points and a maximum of 1000 iterations. This validates the approach with very few iterations but several different well-chosen starting points. 
\subsection{Comparison to other Federated Learning algorithms}
We compare our algorithm to FedMerDel \cite{Rao2025} with greedy search and a federated K-modes algorithm, federated K-means being a very popular federated learning clustering algorithm. We build our federated K-modes algorithm following the federated K-means pipeline proposed by Garst and Reinders (2025) \cite{GarstReinders2025}. We did slightly modify the pipeline by adding a server update at the end of the algorithm and we did set the number of clusters to the ground truth. The results of an MCMC algorithm on the full data are also provided as a gold standard reference for our method. \\
\newline
We compare the considered methods through three aspects: run time, clustering quality, and the point estimates of the model parameters. 

\subsubsection{Run time and clustering quality}
Assessing the run time is quite straightforward. The estimated partition is assessed by comparing it to the true clustering structures using two metrics: the adjusted Rand index \cite{HubertArabie1985} and the variation of information \cite{Meila2003}.  Table \ref{tab:simu_method_comparison_clustering} shows the mean running time in seconds, estimated number of clusters, adjusted Rand index and variation of information (VI) and standard deviations for 100 simulated datasets in different data simulation scenarios. 
\newline

\begin{table}[!ht]
    \centering
    \resizebox{\textwidth}{!}{
    \begin{tabular}{|c|c||c|c|c|c|}
    \cline{3-6}
       \multicolumn{2}{c|}{} &  Fed K-modes & FedMerDel & VCMC & Full MCMC \\
         \hline
         \hline
        \multirow{4}{3cm}{Homogeneous shards easy separability}&  Mean running time in s (sd) & 7 (1) & 12 (18) &  1940 (592) & 4338 (76) \\
         \cline{2-6}
        & Mean number of clusters (sd) & 5 (0) & 5.33 (0.55) &  6.45 (1.14) & 5.19 (0.44) \\
         \cline{2-6}
        & Mean ARI (sd) & 0.70 (0.14) & 0.95 (0.04) & 0.95 (0.03) & 0.96 (0.03)\\
         \cline{2-6}
         & Mean VI (sd) & 0.84 (0.30) & 0.20 (0.12) & 0.18 (0.12) & 0.17 (0.11) \\
         \hline
         \hline
         \multirow{4}{3cm}{Homogeneous shards poor separability} & Mean running time (sd) & 7.74 (1.89) & 6.62 (17.40) & 5174 (3948) & 4668 (379) \\
         \cline{2-6}
         & Mean number of clusters (sd) & 5 (0) & 4.98 (0.32) &  6.02 (1.46) & 5.06 (0.28) \\
         \cline{2-6}
         & Mean ARI (sd) & 0.28 (0.08) & 0.60 (0.10) &  0.61 (0.08) & 0.65 (0.07)\\
         \cline{2-6}
         & Mean VI (sd) & 2.15 (0.23) & 1.32 (0.24) & 1.33 (0.23) & 1.24 (0.20)\\
         \hline
         \hline
         \multirow{4}{3cm}{Heterogeneous nested shards easy separability} & Mean running time (sd) & 9.83 (3.20) & 1.82 (0.14) &  2365 (700) & 4429 (201) \\
         \cline{2-6}
         & Mean number of clusters (sd) & 5.35 (0.94) & 6.18 (0.41) & 7.4 (1.39) & 6.21 (0.41) \\
         \cline{2-6}
         & Mean ARI (sd) & 0.63 (0.12) & 0.97 (0.04) &  0.97 (0.02) & 0.97 (0.03)\\
         \cline{2-6}
         & Mean VI (sd) & 0.99 (0.26) & 0.14 (0.09) & 0.13 (0.08) & 0.16 (0.09)\\
         \hline
         \hline
         \multirow{4}{3cm}{Heterogeneous nested shards poor separability}  & Mean running time (sd) & 10.84 (12.19) & 2.29 (0.32) & 5041 (3494) & 4477 (136) \\
         \cline{2-6}
         & Mean number of clusters (sd) & 6(0) & 5.86 (0.43) & 7.04 & 5.97 (0.36)\\
         \cline{2-6}
         & Mean ARI (sd) & 0.24 (0.09) & 0.67 (0.12) &  0.72 (0.08) & 0.67 (0.08)\\
         \cline{2-6}
         & Mean VI (sd) & 2.23 (0.25) & 1.15 (0.26) & 1.08 (0.22) & 1.23 (0.21)\\
         \hline
         \hline
         \multirow{4}{3cm}{Heterogeneous non-nested shards easy separability} & Mean running time (sd) & 9.70 (3.27) & 1.80 (0.11) &  2200 (504) & 4469 (305) \\
         \cline{2-6}
         & Mean number of clusters (sd) & 5.39 (0.91) & 6.27 (0.62) & 7.51 (1.51) & 6.25 (0.52)\\
         \cline{2-6}
         & Mean ARI (sd) & 0.63 (0.14) & 0.97 (0.03) & 0.98 (0.02) & 0.97 (0.02) \\
         \cline{2-6}
         &Mean VI (sd) & 0.94 (0.28) & 0.13 (0.08) & 0.11 (0.07) & 0.15 (0.08)\\
         \hline
         \hline
         \multirow{4}{3cm}{Heterogeneous non-nested shards poor separability} & Mean running time (sd) & 8.79 (2.15) & 2.17 (0.30) & 4390 (2353) & 4488 (123)\\
         \cline{2-6}
         & Mean number of clusters (sd) & 6 (0) & 5.85 (0.43) &  6.86 & 6.07 (0.38) \\
         \cline{2-6}
         & Mean ARI (sd) & 0.31 (0.10) & 0.70 (0.12) & 0.76 (0.06) & 0.69 (0.07)\\
         \cline{2-6}
         & Mean VI (sd) & 2.16 (0.29) & 1.04 (0.27) & 0.95 (0.19) & 1.20 (0.20) \\
         \hline
    \end{tabular}
    }
    \caption{Mean running time in seconds (s), estimated number of clusters, adjusted Rand index (ARI) and variation of information (VI) and standard deviations (sd) for 100 simulated datasets in different data simulation scenarios.}
    \label{tab:simu_method_comparison_clustering}
\end{table}

As expected, FedMerDel and the federated K-means run around 100 times faster than the sampling based methods. VCMC runs on average two times faster than the MCMC on the full data for easily separable clusters. However, the running times of the two sampling-based methods are comparable on average for data carrying high uncertainty in terms of cluster separability. In this case, the time for the VCMC algorithm from the aggregation step to reach convergence is highly variable. VCMC tends to overestimate the number of clusters in the data, even though this number is well estimated on average by the MCMC algorithm run on the full data or by FedMerDel. However, in terms of adjusted Rand index and of variation of information, the partition estimated by VCMC is on average closer to the true data partition than the ones estimated by FedMerDel or the MCMC on the full dataset.\\

\subsubsection{Parameter estimation}
 The estimates of the model parameters are assessed using the empirical relative bias and the 95\% coverage rate. In the case of Federated K-modes, the algorithm being  non-parametric, the true model parameters are compared to the empirical proportion of observations in each estimated cluster and the empirical proportions of each modality for each variable in each estimated cluster. We can then derive a bias-like estimate, but no 95\% coverage. Table \ref{tab:simu_method_pi} shows the empirical relative bias (RB) and 95\% coverage interval (IC95) on $\pi_c$ estimated on 100 simulated datasets in different data simulation scenarios. \\
\begin{table}[!ht]
    \centering
    \resizebox{\textwidth}{!}{
    \begin{tabular}{|c|c||c|c||c|c||c|c||c|c||c|c||c|c|}
    \cline{3-14}
    \multicolumn{2}{c||}{} & \multicolumn{2}{c||}{$\pi_1$} & \multicolumn{2}{c||}{$\pi_2$} & \multicolumn{2}{c||}{$\pi_3$} & \multicolumn{2}{c||}{$\pi_4$} & \multicolumn{2}{c||}{$\pi_5$}  & \multicolumn{2}{c|}{$\pi_6$}\\
    \hline
    Data & Algorithm & RB & IC95 & RB & IC95 & RB & IC95 & RB & IC95 & RB & IC95 & RB & IC95 \\
    \hline
    \hline
    \multirow{4}{2cm}{homogeneous easy} & Fed K-modes & -0.131 & na  &  -0.175 & na &  -0.113  & na & -0.136 & na & -0.189 & na & na & na \\
    \cline{2-14}
    & FedMerDel & 0.008 & 0.89 & 0.012 & 0.85 & 0.005 & 0.95 &  0.005 & 0.90 &  0.001 & 0.86 & na & na \\
    \cline{2-14}
    & VCMC & 0.007 & 0.90 & 0.006 & 0.90& 0.007 & 0.90 & 0.003 & 0.87 & 0.002 & 0.91 & na & na \\
    \cline{2-14}
    & Full MCMC & 0.002 & 0.89 &  0.002 & 0.88 &  0.001 & 0.94 & -0.002 & 0.89 & -0.002 & 0.89 & na & na\\
    \hline
    \hline
    \multirow{4}{2cm}{homogeneous poor}
    & Fed K-modes & -0.192 & na & 0.198 & na & -0.116  & na & -0.250 & na & -0.201 & na & na & na \\
    \cline{2-14}
    & FedMerDel & -0.073 & 0.56 & -0.071 & 0.64 & -0.042 & 0.64 & -0.071 & 0.63 & -0.057 & 0.65 & na & na \\
    \cline{2-14}
    & VCMC & -0.022 & 0.35 & -0.004 & 0.35 & -0.001 & 0.39 & 0.007 & 0.42 & -0.007 & 0.44 & na & na\\
    \cline{2-14}
    & Full MCMC & -0.014 & 0.58 & 0.002 & 0.56 & 0.001 & 0.61 & 0.001 & 0.50 & -0.003 & 0.58 & na & na \\
    \hline
    \hline
    \multirow{4}{2cm}{heterogeneous nested easy} & Fed K-modes & 0.011 & na & -0.031 & na  & -0.281 & na & -1.281 & na &  -1.241 & na & -4.433 & na \\
    \cline{2-14}
    & FedMerDel & 0.009 & 0.94 & 0.008 & 0.90 & 0.034 & 0.85 & -0.015 & 0.93 & -0.013 & 0.93 & -0.056 & 0.93 \\
    \cline{2-14}
    & VCMC & 0.015 & 0.87 & 0.012 & 0.90 & 0.036 & 0.75 & -0.033 & 0.82 & -0.033 & 0.90 & -0.099 & 0.69 \\
    \cline{2-14}
    & Full MCMC & 0.003 & 0.95 & -0.001 & 0.95 & 0.030 & 0.84 & -0.019 & 0.91 & -0.025  & 0.89 & -0.043 & 0.87 \\
    \hline
    \hline
    \multirow{4}{2cm}{heterogeneous nested poor} & 
    Fed K-modes & 0.142 & na & 0.188 & na & -0.152 & na & -1.335 & na & -1.222 & na & -3.875 & na \\
    \cline{2-14}
    & FedMerDel & -0.012 & 0.69 & -0.026 & 0.71 & -0.054 & 0.62 & -0.613 & 0.50 & -0.372 & 0.63 & -1.618 & 0.51 \\
    \cline{2-14}
    & VCMC & 0.025 & 0.53 & 0.015 & 0.49 & 0.021 & 0.56 & -0.062 & 0.43 & -0.012 & 0.35 & -0.198 & 0.44 \\
    \cline{2-14}
    & Full MCMC & -0.042 & 0.49 & -0.045 & 0.45 & 0.003 & 0.55 & 0.032  & 0.48 & 0.025 & 0.55 & -0.855 & 0.29 \\
    \hline
    \hline
    \multirow{4}{2cm}{heterogeneous non-nested easy} & Fed K-modes & -0.075 & na & -0.018 & na & -0.250 & na & -1.043 & na & -0.931 & na & -2.801 & na \\
    \cline{2-14}
    & FedMerDel & 0.015 & 0.91 & 0.014 & 0.94 & -0.008 & 0.87 & 0.013 & 0.93 & 0.005 & 0.93 & -0.001 & 0.95 \\
    \cline{2-14}
    & VCMC & 0.019 & 0.88 & 0.019 & 0.89 & -0.010 & 0.91 & -0.005 & 0.92 & -0.015 & 0.96 & -0.046 & 0.86 \\
    \cline{2-14}
    & Full MCMC & 0.006 & 0.97 & 0.006 & 0.96 & -0.017 & 0.93 & 0.006 & 0.93 & -0.006 & 0.96 & -0.002 & 0.95 \\
    \hline
    \hline
    \multirow{4}{2cm}{heterogeneous non-nested poor} & Fed K-modes & 0.129 & na & 0.124 & na & -0.229 & na & -0.843 & na & -1.045 & na & -2.890 & na\\
    \cline{2-14}
    & FedMerDel & -0.076 & 0.56 & -0.001 & 0.79 & -0.081 & 0.63 & -0.495 & 0.57 & -0.333 & 0.68 & -0.610 & 0.70\\
    \cline{2-14}
    & VCMC & 0.012 & 0.61&  0.008 & 0.55 & -0.004 & 0.53 & -0.021 & 0.48 & 0.019 & 0.53 &  0.009 & 0.63  \\
    \cline{2-14}
    & Full MCMC & -0.031 & 0.54 &-0.033 & 0.60 & -0.021 & 0.54 & 0.039 & 0.48 &  0.048 & 0.54 & -0.027 & 0.33\\
    \hline
    \end{tabular}
    }
    \caption{Empirical relative bias (RB) and 95\% coverage interval (IC95) on $\pi_c$ estimated on 100 simulated datasets in different data simulation scenarios. (na: not applicable)}
    \label{tab:simu_method_pi}
\end{table}
\newline
For all the data simulation scenarios, the implemented federated K-modes algorithm leads to highly biased estimates. None of the other methods leads to clearly better cluster weight estimates in all the data simulation scenarios considered. In the case of a homogeneous distribution of easily separable clusters across the shards, VCMC, FedMerDel and the MCMC on the full data lead to relative bias on the cluster weights less than or equal to 1\% and 95\% coverage rates around 90\%. The 95\% coverage interval is underestimated by all the methods. This behaviour of FedMerDel is expected as it is often reported in the literature that the mean field approximation used in variational inference leads to an underestimation of the posterior variances \cite{Blei2017}. It is also expected in the case of the two sampling-based methods because the post-processing of the MCMC chain reduces the variability of the MCMC chains \cite{Liverani2015}. In the same distributional setting of the clusters across the shards but with poor separability, VCMC leads to smaller relative bias on the cluster weights than FedMerDel. However, the former underestimates the 95\% credibility interval more often than the latter. In the two cases of a heterogeneous distribution across the shards with easily separable clusters, FedMerDel slightly outperforms VCMC in terms of relative bias and 95\% credibility intervals of the cluster weights. For poorly separable clusters with a heterogeneous distributions across the shards, FedMerDel still performs better in terms of assessing 95\% coverage interval but leads to highly biased estimates of the small clusters' weights (with relative bias up to 1618\% in absolute value). It is interesting to note that in this case, VCMC allows better estimates of the small clusters' weights than the MCMC on the full data. For example, in the case of a non-nested distribution of the clusters across the shards, the relative bias on $\pi_6$ is equal to 0.9\%, and the 95\% coverage rate to 63\% with VCMC, whereas those criteria are equal to -2.7\% and 33\% when the inference is performed with the MCMC on the full data. Similar finding can be made for the parameters $\pi_1$ and $\pi_5$ in the same simulation setting and the parameters $\pi_5$ and $\pi_6$ in the case of a heterogeneous nested distribution of the clusters across the shards. These results are explained by the fact that in these settings, the sizes of these clusters are small in the full data, but they are big enough within some shards for VCMC to identify them with less uncertainty than the full MCMC.\\ 
\newline
Table \ref{tab:simu_method_comp_p} shows the empirical relative bias (RB) and 95\% coverage interval (IC95) on $p_{k,c}^q$ estimated on 100 simulated datasets in different data simulation scenarios averaged within six ranges of values of $p_{k,c}^q$. All the studied algorithms lead to highly biased estimates of $p_{k,c}^q$ smaller than 0.1. This is because the size of the datasets is 4000 which is too small to infer such values of $p_{k,c}^q$. FedMerDel leads to estimates on the parameters $p_{k,c}^q$ with smaller relative bias than VCMC. However, the former underestimates massively the uncertainty on these parameters with 95\% credibility intervals always smaller than 13\%. Once again, it is interesting to note that for heterogeneous distribution of the clusters across the shards, VCMC tends to perform better than the MCMC on the full data in terms of estimating the model parameters related to small clusters. 

\begin{table}[!ht]
    \centering
    \resizebox{\textwidth}{!}{
    \begin{tabular}{|c|c||c|c||c|c||c|c||c|c||c|c||c|c|}
    \cline{3-14}
    \multicolumn{2}{c||}{} & \multicolumn{2}{c||}{$p_{k,c}^q< 0.0001$} & \multicolumn{2}{c||}{$p_{k,c}^q\in [0.0001, 0.1[$} & \multicolumn{2}{c||}{$p_{k,c}^q \in [0.1, 0.5[$} & \multicolumn{2}{c||}{$p_{k,c}^q \in [0.5, 0.9[$} & \multicolumn{2}{c||}{$p_{k,c}^q \in [0.9, 0.9999[$} & \multicolumn{2}{c|}{$p_{k,c}^q \geq 0.9999$}  \\
    \hline
    Data & Algorithm & RB & IC95 & RB & IC95 & RB & IC95 & RB & IC95 & RB & IC95 & RB & IC95 \\
    \hline
    \hline
    \multirow{4}{2cm}{homogeneous easy} & Fed K-modes & -2e7 & na & -40.154 & na &  0.011 & na & 0.020 & na & 0.069 & na & 0.076 & na \\
    \cline{2-14}
    & FedMerDel & -2e7 & 0.05 & -0.439 & 0.10 & 0.001 & 0.06 & -0.001 & 0.07 &  0.001 & 0.11 & 0.001 & 0.11 \\
    \cline{2-14}
    & VCMC & -9e7 & 0.14 & -6.54 & 0.55 & -0.012 & 0.87 & -0.010 & 0.83 & 0.013 & 0.51 & 0.012 & 0.23 \\
    \cline{2-14}
    & Full & -2e7 & 0.25 & -2.82 & 0.76 & -0.010 & 0.90 & 0.006 & 0.88 & 0.006 & 0.73 & 0.006 & 0.33 \\
    \hline
    \hline
    \multirow{4}{2cm}{homogeneous hard} & Fed K-modes & -2e4 & na & -5.869 & na & -0.052 & na & 0.045 & na & 0.126 & na & na & na \\
    \cline{2-14}
    & FedMerDel & -7e2 & 0.00 & -0.509 & 0.04 & -0.012 & 0.07 & 0.009 & 0.10 & 0.008 & 0.08 & na & na \\
    \cline{2-14}
    & VCMC & -1e4 & 0.00 & -3.902 & 0.22 & -0.081 & 0.64 & 0.058 & 0.48 & 0.085 & 0.18 & na & na \\
    \cline{2-14}
    & Full & -6e3 & 0.00 & -2.620 & 0.23 & -0.075 & 0.58 & 0.051 & 0.42 & 0.057 & 0.18 & na & na \\
    \hline
    \hline
    \multirow{4}{2cm}{heterogeneous nested easy} & Fed K-modes & -9e10 & na & -63.391 & na &-0.047 & na & 0.062 & na & 0.120 & na & 0.143 & na \\
    \cline{2-14}
    & FedMerDel & -5e9 & 0.06 & -2.800 & 0.11 & -0.001 & 0.07 & 0.001 & 0.07 & 0.004 & 0.13 & 0.004 &0.11 \\
    \cline{2-14}
    & VCMC & -4e10 & 0.16 & -4.856 & 0.64 & -0.012 & 0.89 & 0.009 & 0.87 & 0.010 & 0.61 & 0.010 & 0.25 \\
    \cline{2-14}
    & Full & -1e9 & 0.22 & -3.345 & 0.74 & -0.011 & 0.89 & 0.007 & 0.87 & 0.008 & 0.72 & 0.007 & 0.33 \\
    \hline
    \hline
    \multirow{4}{2cm}{heterogeneous nested poor} 
    & Fed K-modes & -8e2 & na & -9.902 & na & -0.116 & na & 0.093 & na & 0.158 & na & na & na \\
    \cline{2-14}
    & FedMerDel & -14.546 & 0.00 & -1.184 & 0.05 & -0.041 & 0.09 & 0.029 & 0.10 & 0.023 & 0.10 & na & na\\
    \cline{2-14}
    & VCMC & -1e2 & 0.00 & -3.394 & 0.32 & -0.070 & 0.71 & 0.051 & 0.59 & 0.059 & 0.29 & na & na \\
    \cline{2-14}
    & Full & -2e2 & 0.00 & -3.512 & 0.24 & -0.079 & 0.61 & 0.056 & 0.46 & 0.062 & 0.24 & na & na \\
    \hline
    \hline
    \multirow{4}{2cm}{heterogeneous non-nested easy} & Fed K-modes & -2e8 & na & -74.614 & na & -0.080 & na & 0.0778 & na & 0.150 & na & 0.153 & na \\
    \cline{2-14}
    & FedMerDel & -6e6 & 0.05 & -3.65 & 0.08 & -0.004 & 0.06 & 0.003 & 0.07 & 0.003 & 0.12 & 0.008 & 0.12 \\
    \cline{2-14}
    & VCMC & -4e7 & 0.18 & -4.606 & 0.65 & -0.011 & 0.89 & 0.008 & 0.87 & 0.009 & 0.62 & 0.008 & 0.24\\
    \cline{2-14}
    & Full & -2e7 & 0.26 & -5.656 & 0.75 & -0.019 & 0.88 & 0.011 & 0.87 & 0.011 & 0.73 & 0.006 & 0.33 \\
    \hline
    \hline
    \multirow{4}{2cm}{heterogeneous non-nested poor} &  Fed K-modes & -1e4 & na & -9.244 & na &  -0.011 & na & 0.086 & na & 0.1714 & na & na & na\\
    \cline{2-14}
    & FedMerDel &  -1e2 & 0.00 & -1.058 & 0.05 & -0.041 & 0.07 & 0.027 & 0.09 & 0.021 & 0.08 & na & na\\
    \cline{2-14}
    & VCMC & -1e3 & 0.00 & -3.043 & 0.34 & -0.058 & 0.75 & 0.041 & 0.61 & 0.049 & 0.33 & na & na \\
    \cline{2-14}
    & Full &  -2e3 & 0.00 & -3.325 & 0.23 & -0.078 & 0.63 & 0.053 & 0.45 & 0.056 & 0.24 & na & na \\
    \hline
    \end{tabular}
    }
    \caption{Empirical relative bias (RB) and 95\% coverage interval (IC95) on $p_{k,c}^q$ estimated on 100 simulated datasets in different data simulation scenarios averaged within six ranges of values of $p_{k,c}^q$. (na: not applicable)}
    \label{tab:simu_method_comp_p}
\end{table}
\noindent
\section{Application to electronic health record (EHR) data: \newline Multi-morbidity in the THIN cohort }
\label{sec:application_on_THINdata}
We apply our algorithm to identify common patterns of long-term disease co-occurrence in the UK geriatric population. We use a subset of The Health Improvement Network (THIN) database \cite{BourkeDattaniRobinson2004, Lewis2007}. This database is composed of EHRs collected at the primary care level throughout the UK. We select individuals over 80 years old who were reported to suffer from at least two of the long term health conditions listed in Appendix \ref{app:THIN} between 1\textsuperscript{st} January 2010 and 1\textsuperscript{st} January 2017. This subset contains 289,821 individuals. \\

\begin{table}[!ht]
    \centering
    \begin{tabular}{|rc|}
         \hline
         \multicolumn{1}{|l}{\textbf{Mortality} (\%)} &48.1 \\
        \hline
        \multicolumn{2}{|l|}{\textbf{Sex} (\%)} \\
        Male & 35.3 \\
        Female & 64.7 \\
        \hline
        \multicolumn{2}{|l|}{\textbf{Age at entry}(in years) } \\
        Mean (sd) & 85.4 (4.4) \\
        Min - Max & 80 - 112\\
        \hline
        \multicolumn{2}{|l|}{\textbf{Age at death }(in years)}\\
        Mean (sd) & 89.4 (4.7) \\
        Min - Max & 80 - 110 \\
        \hline
        \multicolumn{2}{|l|}{\textbf{Region of residency} (\%)}\\
         East Midlands & 2.4 \\
        East of England & 5.7 \\
        London & 10.4 \\
        North East & 1.8 \\
        North West & 9.2 \\
        Northern Ireland & 3.9 \\
        Scotland & 14.9 \\
        South Central & 10.4 \\
        South East Coast & 11.0 \\
        South West & 8.7 \\
        Wales & 12.0 \\
        West Midlands & 8.0 \\
        Yorkshire \& Humber & 1.6 \\
        \hline
        \multicolumn{2}{|l|}{\textbf{Number of co-occurring diseases}}\\
         Mean (sd)& 6.2 (3.1) \\
         Min - Max & 2 - 33\\
         \hline
    \end{tabular}
    \caption{Description of the studied subset of the THIN data (N = 289,821)}
    \label{tab:THINsummary}
\end{table}
\noindent
Table \ref{tab:THINsummary} provides information on the studied population.  64.7\% of the individuals are female. The repartition of the population over the country is not representative of the distribution of the general population across the UK in 2011, with a higher representation of residents from Wales and Scotland and a low representation of residents from Yorkshire \& Humber. The mean age at entry in the study is 85.4 years with a range lying between 80 and 112 years. The mean age at death is 89.4 with a range lying between 80 and 110. The mean number of co-occurring diseases is 6.2 with a range of value lying between 2 and 33.\\

\begin{table}[!ht]
    \centering
    \begin{tabular}{|c|c||c|c||c|c|}
      \hline
    Cluster & N & Cluster & N & Cluster & N \\ 
      \hline
      1 & 138,607 &  10 & 5,207 &  19 &  89 \\  
        2 & 3,687 &  11 & 2,136 & 20 & 215 \\ 
        3 & 11,332 & 12 & 2,404&  21 &  31 \\  
        4 & 22,410 &  13 & 3,435 & 22 &  61 \\  
        5 & 28,395 &  14 & 3,205 & 23 & 478 \\ 
        6 & 12,080 & 15 & 100 & 24 & 233 \\   
        7 & 28,078 & 16 & 1,007 &  25 & 259 \\   
        8 & 3,830 &  17 & 263 &  26 &  92 \\ 
        9 & 21,121 & 18 & 1,000 & 27 &  66 \\  
       \hline
    \end{tabular}
    \caption{Number of individuals (N) per cluster identified in the studied population (N = 289,821).}
    \label{tab:THIN_n_indiv_clust}
\end{table}

\begin{figure}[!ht]
    \centering
    \includegraphics[width=1\linewidth]{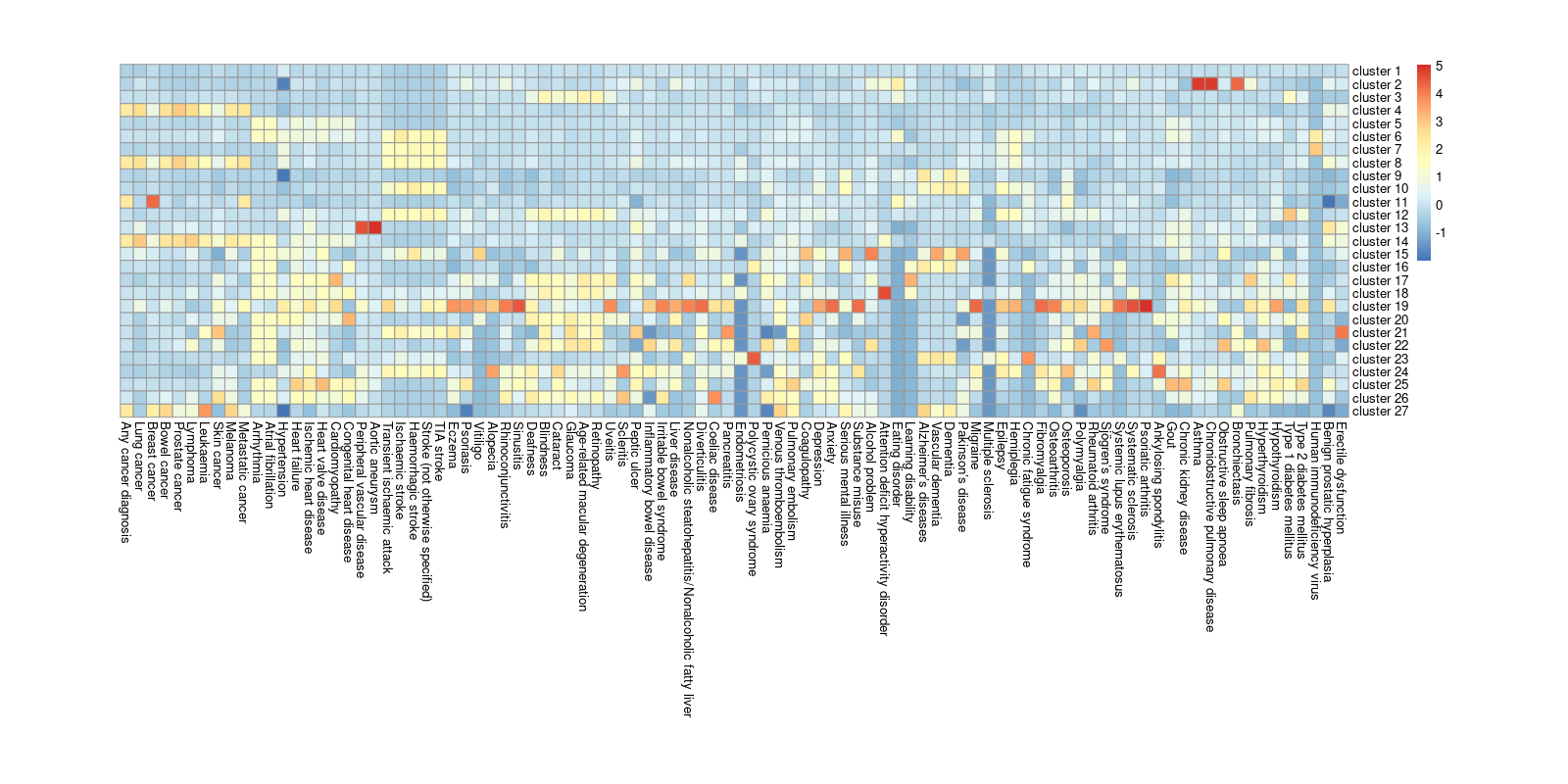}
    \caption{Scaled difference between the proportion of individuals presenting each disease in each cluster and the proportions in the full data}
    \label{fig:THIN_clusters_summary}
\end{figure}

\noindent
The VCMC framework with the Ball matching is applied to the data. The data are split into 30 shards with a random uniform shard allocation. 27 clusters are identified in the studied population. Table \ref{tab:THIN_n_indiv_clust} shows the number of individuals per cluster. Cluster 1 comprises 138,607, representing 47.83\% of the studied population. The smallest cluster identified is cluster 21, with 31 individuals (0.01\% of the studied population). Figure \ref{fig:THIN_clusters_summary} shows the scaled difference between the proportion of individuals presenting each disease in each cluster and the proportions in the full data. The proportion of individuals presenting each disease in each cluster is shown in Appendix \ref{app:THIN}. Table \ref{tab:THIN_clusters_characterisation} summarises this figure by characterising in each cluster the conditions with higher/lower prevalence compared to the full studied population. Cluster 1 is not characterised by any specific condition and is representative of the total studied population. Cluster 7 is the second largest cluster. It is characterised by a high prevalence of stroke. Interestingly, a high proportion of patients infected by HIV are in this cluster, indicating that HIV is not associated with any specific co-morbidity and then no cluster specific to HIV patients is created. Clusters 3, 4, 5, 6, 7 and 9 are of significant size. They are respectively characterised by high prevalences of deafness and ophthalmic disorders, cancers, chronic heart disorders, chronic heart disorders and stroke and dementia. All of those diseases are very common across the UK geriatric population. Cluster 11, being characterised by a high breast cancer prevalence and a low prevalence of benign prostatic hyperplasia, is probably composed of females. It is also the only cluster that gathers patients with a specific cancer sub-type among the five clusters (4, 8, 11, 14, 27) that stand out for gathering individuals suffering from cancer. Cluster 19 is a small cluster, gathering only 89 individuals, but is characterised by a high multi-morbidity. Cluster 21, which is the smallest cluster, is characterised by pancreatitis, rheumatoid arthritis and erectile dysfunction. Further details on clusters characterisation are provided Appendix \ref{app:THIN_cluster}.

\begin{sidewaystable}[]
    \centering
    \resizebox{\textwidth}{!}{
    \begin{tabular}{|c|l|l|}
    \hline
       Cluster  &  Low condition prevalence & High condition prevalence \\
       \hline
       1  & \multicolumn{2}{c|}{No specific condition} \\
       \hline
       2 & Hypertension & Asthma, Chronic obstructive pulmonary disease, Bronchiectasis\\
       \hline
       3 & & Deafness, Ophthalmic disorders\\
       \hline
       4 & & Cancers \\
       \hline
       5 & & Chronic cardiac disorders \\
       \hline
       6 & & Chronic cardiac disorders, stroke\\
       \hline
       7 & & Stroke, HIV \\
       \hline
       8 & & Cancers, chronic cardiac disorders \\
       \hline
       9 & Hypertension & Dementia \\
       \hline
       10 & & Stroke, Dementia \\
       \hline
       11 & Benign prostatic hyperplasia & Breast cancer, Metastatic cancer \\
       \hline
       12 & & Stroke, Deafness, Ophthalmic disorders, Type-1 diabetes\\
       \hline
       13 & & Peripheral vascular disease, Aortic aneurysm, Benign prostatic hyperplasia\\
       \hline
       14 & & Cancer, Chronic cardiac disorders, Benign prostatic hyperplasia \\
       \hline
       15 & Endometriosis, Multiple sclerosis & Alcohol problems, Serious mental illness, Vascular dementia, Parkinson's disease, \\
       &  & Chronic cardiac disorder, Stroke, Vitiligo, Obstructive sleep apnoea \\
       \hline
       16 & & Polycystic ovary syndrome, Dementia, Arrhythmia, Atrial fibrillation, Chronic fatigue syndrome \\
       \hline
       17 & Endometriosis & Cardiomyopathy and other chronic cardiac disorders, Deafness, Ophthalmic disorders, \\
       & & Learning disability, Pulmonary fibrosis\\
       \hline
       18 & & ADHD, Chronic cardiac disorders, Ophthalmic disorder\\
       \hline
       19 & Endometriosis & Substance misuse, Sinusitis, Rhinoconjunctivitis, Dermatological disorders, Uveitis, \\
       && Gastrointestinal disorders, Depression, Anxiety, Migraine, Autoimmune disorders, \\
       && Skeletal disorders, Neurological disorders, Fibromyalgia\\
       \hline
       20 & & Congenital heart disease and other chronic cardiac disorders, Deafness, Ophthalmic disorder,\\
       &&Coagulopathy \\ 
       \hline
       21 & Endometriosis, Parkinson's disease & Pancreatitis, Peptic ulcer, Erectile dysfunction\\
       \hline
       22 & & Sj{\"o}gren’s syndrome,  Polymyalgia, Thyroid disorders\\
       \hline
       23 & & Polycystic ovary syndrome, Chronic fatigue disorder\\
       \hline
       24 & & Ankylosing spondylitis, Alopecia, Scleritis\\
       \hline
       25 & & Heart valve disease, Heart failure and other chronic cardiac disorder, Gout,  \\ && Chronic kidney disease, Type-2 diabetes mellitus\\
       \hline
       26 & Inflammatory bowel disease & Coeliac disease, Pernicious anaemia, Scleritis\\
       \hline
       27 & Hypertension, Psoriasis, Pernicious anaemia, Benign prostatic hyperplasia & Breast cancer, Leukaemia \\
       \hline
    \end{tabular}
    }
    \caption{Characterisation of the clusters identified in the studied population}
    \label{tab:THIN_clusters_characterisation}
\end{sidewaystable}

\section{Discussion}
\label{sec:discussion}
In this work, we presented a comprehensive framework to perform VCMC in order to infer Bayesian mixture models in a Federated Learning setting. The method is a Consensus Monte Carlo approach and its originality lies in the way of combining the local chains. We extended the method, first proposed by Rabinovich, Angelino and Jordan (2015) \cite{rabinovich2015}, to allow the inference of all the parameters of an over-fitted mixture model. Although we focused on over-fitted mixtures of categorical distributions, the framework can be directly adapted to finite mixture models by restricting the number of clusters or mixtures of other (even complex) distributions. We also provided different inference strategies that can adapt to several federated learning settings. \\
\newline
\subsection{Main findings} 
\subsubsection{Cluster matching}
A major challenge of {\em Consensus Monte Carlo} for unsupervised clustering and classification is matching the local clusters across the shards. Rabinovich, Angelino and Jordan (2015) \cite{rabinovich2015} used the Hungarian algorithm to match the clusters based on the location of the cluster-specific distributions. However, this method is always failing when there is no shard where all clusters are inferred. This condition cannot be assumed in a cross-silo federated learning setting, which was the focus of our study. We provided two different ways of matching the clusters across the shards: the minimum divergence matching and the Ball matching. We showed in the simulation study that the minimum divergence matching performs worse than the Ball matching in terms of estimating the number of clusters. This is due to the fact that the minimum divergence matching cannot match clusters that are estimated within a same shard whereas the Ball matching can. By doing so, the Ball matching is able to mitigate any overestimation of the number of clusters that occurs at the local level. However, we also showed that the value of the adjusted Rand index between the estimated partition and the true partition is on average the same with the two matching indicating that the minimum divergence matching leads to some existing clusters being divided into sub-clusters. On the contrary, the Ball matching sometimes merges clusters that should not be merged. The choice of the matching algorithm should depend on the application. If estimating the number of clusters is an important stake then Ball merging should be preferred. However, if this problem is not the main focus, but abusive cluster merging should absolutely be avoided, then the minimum divergence matching stands as the best option. 
\subsubsection{Performances of VCMC for mixture models}
We assessed the performance of our method and compared it to other inference methods to perform unsupervised clustering and classification in Federated Learning settings. 
\paragraph{Comparison to Federated K-modes} federated K-modes is the only non-Bayesian method we considered. It is an adaptation of the federated K-means algorithm proposed by Garst and Reinders (2025) \cite{GarstReinders2025}. This method was selected because of the popularity of the K-means algorithm for clustering tasks and its simplicity of implementation. Despite the fact that we set the number of clusters to the truth, the performance was disappointing in terms of estimating the clustering structure of the data, and therefore the estimates of the proportions of observations per clusters and the proportions of the modalities of each of the categorical variables in each cluster. We acknowledge that the original method was developed for a clustering task of continuous variables. In this setting, the distance between points is the Euclidean distance which is also continuous and the centres of each cluster are the centroids of the data points in the cluster. In a K-modes setting, the distance between observation is the minimum mismatch, which is finite and the centres of the clusters are the majority mode. Some further adjustment to account for these differences might be necessary to develop a competitive federated learning K-modes algorithm.
\paragraph{Comparison to FedMerDel and full MCMC} We also compared VCMC to FedMerDel \cite{Rao2025}, a federated variational inference setting for over-fitted mixture of categorical distributions. To our knowledge, it is the only existing federated learning method that assumes the exact same statistical models as us. No clear winner appeared between VCMC and FedMerDel based on the comparison criteria we studied. The main strength of FedMerDel is its speed whereas VCMC is, in some settings, not faster than a MCMC algorithm on the full data. This slowness can be mitigated by a better choice of the step size in the gradient descent algorithm of the aggregate step. Indeed, this parameter must be tuned carefully because a value that is too small leads to very slow convergence of the algorithm, but a value that is too large leads to steps that overshoot the probability simplex, which hinders the identification of the optimum within this constrained space. When the distribution of the clusters across the shard is heterogeneous and the clusters are poorly separable, FedMerDel struggled to identify small clusters, even if they represent a significant part of the population within one shard. In contrast, VCMC performed well in this setting, and even better than the MCMC on the full data. Although not our primary focus, we showed that VCMC can be a better option than the MCMC on the full data to infer a mixture model when there exists a strong influence of the data collection pattern (for example, geographical or temporal) on the clustering structure. In this case, using the VCMC framework allows to identify clusters that are small in the global population but of significant size within at least one of the data shards, and leads to better estimates of the marginal posterior distributions of the model parameters related to these clusters.
\paragraph{Comment on the simulation setting} The setting with poorly separable clusters is, by design, leading more often to failure of the unsupervised classification algorithm. However, we believe that this setting is the most comparable to real data applications. Indeed, in the vast majority of the health-related applications of unsupervised clustering and classification algorithms on real data, data are not easily separable into a partition. This behaviour can be partly explained by two facts: 1) the unsupervised clustering task is used as an exploratory analysis to identify subgroups of a population that have similar characteristics, the assumption that there is a clear clustering structure in the data is wrong but convenient; 2) there exists an underlying clustering structure in the data generation mechanism, but the observed variables are capturing this structure only partly. 

\subsection{Application to EHR data}
We showed the interest of VCMC for real data application by identifying patterns of multi-morbidity in the UK geriatric population based on electronic health records. Rao et al. (2025) \cite{Rao2025} applied FedMerDel to identify a clustering structure in this same data. They identified 8 or 17 clusters depending on the version of the algorithm they used. We identified more clusters, 27 in total. However, the difference comes mainly from the fact that we identify some small extra clusters. Indeed, as in Rao et al. (2025) \cite{Rao2025} the biggest cluster identified, containing almost 50\% of the population, was not associated with any specific multi-morbidity pattern. In both studies, the occurrence of cancers, arrhythmia and atrial fibrillation, stroke, ophthalmic-related conditions and dementia seemed to influence strongly the identified clustering structure. Observations arising from our identified clustering structure can be explained by the scientific literature. Clusters where a low prevalence of endometriosis is found were often characterised by conditions that have higher prevalence among males such as cardiomyopathy \cite{Bergan2025}, substance misuse \cite{Maxineau2025} and alcohol problems \cite{NHSEngland2026HSE}. The smallest cluster we identified, containing 31 individuals, is characterised by pancreatitis, rheumatoid arthritis and erectile dysfunction. The correlation between these three conditions is well documented in the medical literature \cite{Chang2015, Wilton2021, Shah2024}. Cluster 27 showed high prevalence of cancers and leukaemia and low prevalence of pernicious anaemia. The strong correlation between cancer and anaemia being well known \cite{Ludwig2004}, patients suffering from cancer are more likely to be B12 supplemented. We observed multi-morbidity patterns with hypertension that seems at first glance at odds with what is reported in the literature. Indeed, cluster 2 and 9 had both a low prevalence of hypertension and respectively a high prevalence of lung-related conditions and of dementia. A strong positive correlation between those two conditions and hypertension is reported in the literature \cite{SPRINT2019, NAthan2019, Livingston2024, Lekamge2026}. However hypertension being so well identified in the medical community as an aggravating factor for these two conditions, it is possible that hypertension is taken into account as part of the global treatment against one of those two conditions and is not reported as a condition in itself by general practitioners. We showed that our identified clustering structure makes sense clinically. However, some of the observed difference in condition prevalences in the identified clusters might be due to randomness. Furthermore, the EHR data being collected at the general practice level, the definition of conditions might vary slightly from one practice to another. For these two reasons, we do not believe that results from this type of study should be used to draw final conclusions on multi-morbidity patterns. However, this type of study is a cost-effective and straightforward way to screen a large population and identify tendencies that could steer further epidemiological investigations.

\subsection{Limitations and future work}
\subsubsection{Cluster matching}
\paragraph{Minimum divergence matching}
The minimum divergence matching minimises the posterior expectation of the Kullback Leibler divergence between the local distributions of the cluster membership. In order to perform this step, a subset of the dataset, called anchor points, can be exposed to the global node, in the spirit of the work of Ni, Ji and Müller (2020) \cite{Ni2020b}. However, we chose to stay in a fully federated learning setting where none of the data are shared between nodes. To do so, we start by selecting one of the local posterior distributions as the reference distribution. Then, we iteratively match the clusters from this reference distribution to the clusters from the other local posterior distributions and update the reference distribution according to the identified matching. Because the process is iterative, the output might depend on the chosen starting point and on the order in which the local distributions are considered for matching. If one shard is known to contain more information than the others (for example it contains more observations or a distribution of the clusters that is representative of the global distribution), then we would recommend using the local posterior distribution inferred in this shard as starting point for the reference distribution. Otherwise, we would recommend performing the matching several times, with several starting points and permutation of the order of the local posterior distributions and selecting the run that minimises the sum of the divergence estimated at each iteration of the matching algorithm. Furthermore, the update of the reference posterior is performed by averaging the matched local posteriors. No additional information, including how representative the shards are of the full data, is taken into account. The averaging could be performed in a Consensus Monte Carlo style, by considering a weighted mean of the local estimates with weights inversely proportional to the uncertainty on these estimates. Another alternative would be to perform the VCMC framework a first time to estimate aggregation weights, then reperform the matching with these first aggregation weights and the inference of the aggregation weights to refine the results.
\paragraph{Ball matching}
The Ball matching groups clusters that have similar point estimates of the parameters of the cluster-specific distributions. We defined close clusters based on the convergence results of Ho and Nguyen (2016) \cite{HoNguyen2016}. However, in their work, a uniform prior distribution is assumed on the parameters of the cluster-specific distributions. This assumption is met in our model if $\beta^q$ is set to 1 for all $q$, but in our application we set $\beta^q$ to $\frac{1}{2}$ as it defines the Jeffreys prior. Despite this flaw, the Ball matching behaved correctly in our simulation and real data application, but we would encourage the user of our framework to update the definition of close clusters if they are aware of any better suited theoretical results on the convergence rate of over-fitted mixture models. If the VCMC framework is applied to a finite mixture model, the rate of convergence should be changed to $\big(\frac{\log n}{n}\big)^{1/2}$ to remain compliant with the work of  Nguyen (2013) \cite{Nguyen2013} and Ho and Nguyen (2016) \cite{HoNguyen2016}. The Ball matching performed better in assessing the number of clusters in the matching comparison study than in the comparison with other methods study. The datasets being the same in those two settings, the difference lies in the number of MCMC iterations used to assess the local posterior distributions. Indeed, a thinning interval of size 5 was applied to the local MCMC chains in the second study. Because the distance between the different clusters is calculated based on the median of the marginal posterior distributions of the parameters of the cluster-specific distributions, a lack of iterations can lead to poor estimates of these medians. A way to mitigate this problem would be to use a distance between the marginal posterior distributions of the parameters, and not between some point estimates derived from these distributions. However, we are not aware of any theoretical result on the convergence rate of those and then, the threshold to merge clusters would have to be determined arbitrarily.
\subsubsection{Shards definition}
As demonstrated in the simulation study, the distribution of the data into the shards has an influence on the estimation of the partition of the data, especially the ability of the algorithm to discover small clusters. This behaviour of the inference algorithm indicates that a minimum proportion of individuals belonging to a given cluster is needed to be able to identify this cluster with a MCMC algorithm. Even though this problem was out of the scope of our study, it would be interesting to investigate this minimum proportion through practical and theoretical studies. The existence of this minimum proportion also means that the number of shards has an impact on the estimated partition. We think that it is likely that the more shards there are, the more clusters will be estimated. Once again, more investigations are needed.
\subsubsection{Extending the results to other types of mixture models}
Even though we claim that the VCMC pipeline we propose is extendable to other types of mixture models, we acknowledge that the performance of our model was only studied for mixture of categorical distributions. Some of the findings from the simulation study might not be generalisable to other types of distribution. It would be interesting to perform the simulation studies we proposed with mixture of Gaussian distributions and mixture of more complex distributions to assess if our findings are consistent across different choices of distributions. 
\subsubsection{Prior fractionation}
We noticed that prior fractionation is often neglected if not totally missing in the literature of Bayesian inference for unsupervised clustering. Ignoring the fractionation of the prior distribution at the local level leads to the definition of a prior distribution at the global level that is much more informative than intended. This reduces the variance of the samples drawn in the local and global posterior distributions and so the assessment of the uncertainty on the model parameters' estimates. We studied briefly the impact of prior fractionation through a simulation study (results in Appendix \ref{app:prior_fractionation}). From our first observations, the prior fractionation seems to have a limited impact on the estimated partition when considering over-fitted mixture models. This is probably not the case for Dirichlet Process mixture models as the concentration parameter $a$ influences the probability of creating a new cluster. In this case, it is straightforward to show that for $a \neq 1$ the probability of creating a new cluster differs with or without prior fractionation and, more precisely, for $a < 1$ (resp. $a >1$), the probability of assigning an individual to an empty cluster is higher (resp. lower) with prior fractionation than without. For these reasons, we think that more practical and theoretical investigations should be performed on the topic.

\bibliographystyle{unsrt}
\bibliography{sample}

\appendix

\section{Comparison of the cluster matchings - removing the impact of the local partition estimation} 
\label{app:cluster_matching}
A comparison of the matching algorithm is provided Section \ref{sec:matching_comparison}. In this study, clusters are estimated locally in each of the shards before being matched.
The recovery of the correct matching depends on the quality of the partitions estimated in the local shards. This remark motivates a second study for which the estimated partitions in each shard is set to the truth. This allows to assess only the matching and not a mixture between the partition's estimation at the local level and the matching at the global level.\\
\begin{table}[]
    \centering
    \resizebox{\textwidth}{!}{
    \begin{tabular}{|c|c|c|c|c|}
    \hline
        Type of shards & Cluster separability & Matching methods & Correct matching recovered & mean ARI \\
        \hline
         \multirow{6}{*}{homogeneous}& \multirow{3}{*}{easy} & Hungarian & 100\% & 1.00 \\
         \cline{3-5}
         & & minimum divergence & 100\% & 1.00 \\
         \cline{3-5}
         & & Ball & 100\% & 1.00\\
         \cline{2-5}
         & \multirow{3}{*}{poor}  & Hungarian & 100\% & 1.00 \\
         \cline{3-5}
         & & minimum divergence & 100\% & 1.00 \\
         \cline{3-5}
         & & Ball & 96\% & 0.99\\
         \hline
         \multirow{6}{*}{heterogeneous nested}& \multirow{3}{*}{easy} & Hungarian & 100\% & 1.00 \\
         \cline{3-5}
         & & minimum divergence &  94\% & 0.99 \\
         \cline{3-5}
         & & Ball & 100\%  & 1.00 \\
         \cline{2-5}
         & \multirow{3}{*}{poor}  & Hungarian & 100\% & 1.00\\
         \cline{3-5}
         & & minimum divergence & 93\% & 0.99\\
         \cline{3-5}
         & & Ball & 90\% & 0.99 \\
         \hline
         \multirow{6}{*}{heterogeneous non-nested}& \multirow{3}{*}{easy} & Hungarian & 0\% & 0.93 \\
         \cline{3-5}
         & & minimum divergence & 99\% & 1.00 \\
         \cline{3-5}
         & & Ball  & 100\%  & 1.00\\
         \cline{2-5}
         & \multirow{3}{*}{poor}  & Hungarian & 0\% & 0.92\\
         \cline{3-5}
         & & minimum divergence & 86\% & 0.99 \\
         \cline{3-5}
         & & Ball & 93\% & 0.99 \\
         \hline
    \end{tabular}
    }
    \caption{Percentage of simulation scenarios for which the correct matching is recovered and mean adjusted Rand index between the estimated partition and the true partition when the local partitions are set to the truth for 100 simulated datasets under the different scenarios.}
    \label{tab:matching_comparison_true_C}
\end{table}

\begin{figure}
    \centering
    \includegraphics[width=0.8\linewidth]{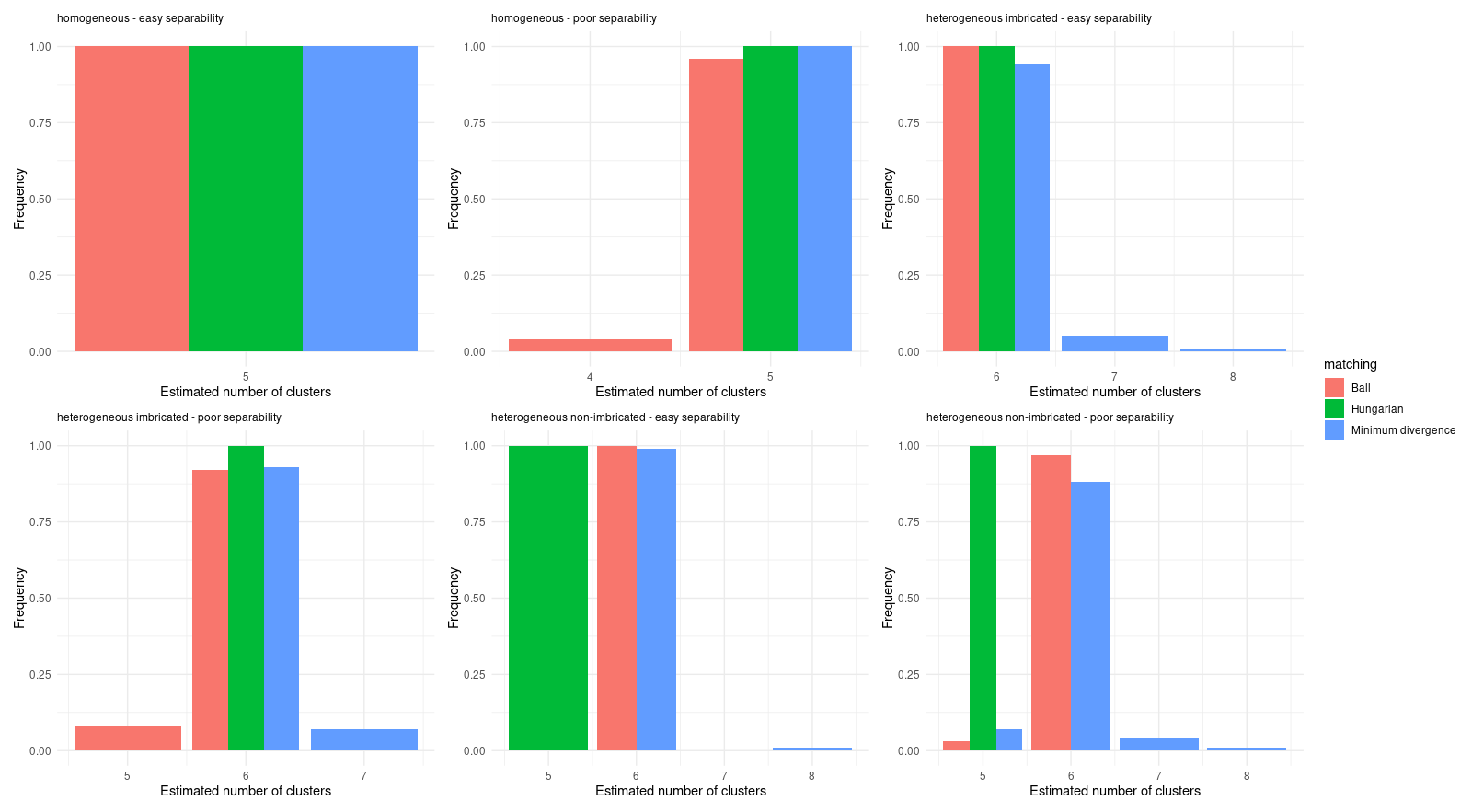}
    \caption{Distribution of the estimated global number of clusters over 100 datasets for local clusters being set to the truth for different matching algorithm and different clusters repartition in the shards.}
\label{fig:barplot_n_clust_matching_comparison_trueC}
\end{figure}
\noindent
Table \ref{tab:matching_comparison_true_C} shows the percentage of simulation scenarios for which the correct matching is recovered and the mean adjusted Rand index between the estimated partition of the full data and the true partition when the local partitions are set to the truth for 100 simulated datasets under the different scenarios. \\
Table \ref{tab:matching_comparison_true_C} shows the percentage of simulation scenarios for which the correct matching is recovered and the mean adjusted Rand index between the estimated partition of the full dataset in the true partition when the partition in the local shards is set to the truth for 100 datasets. In the case of easily separable clusters that are homogeneously distributed across shards, the three matching algorithms always recover the correct matching. For poorly separable clusters still homogeneously distributed across shards, the Hungarian matching and the minimum divergence matching always recovers the true matching whereas the Ball matching fails to recover it in 4\% of the simulations. In the case of easily separable clusters that have a heterogeneous nested distribution across the shards, the Hungarian matching and the Ball algorithms always recover the correct matching, whereas the minimum divergence matching fails to recover the true matching in 6\% of the simulations. For poorly separable clusters with a heterogeneous nested distribution across shards, the Hungarian matching always recovers the correct matching, whereas the minimum divergence matching fails to recover the true matching in 7\% of the simulations and the Ball matching in 10\% of the simulations. In the case of a non-nested heterogeneous distribution of the clusters across the shards, the Hungarian matching always fails to recover the true matching for easily and poorly separable clusters. Once again, this result is expected and is due to the inability of the Hungarian matching to not provide a one to one correspondence for as many clusters as possible. In this setting for clusters distribution across shards, the minimum divergence and the Ball matching algorithms perform well for easily separable clusters with respectively 1\% and 0\% failure in recovering the true matching. The performance of the two matching algorithms decreases in the case of poorly separable clusters. In this case, the Ball matching performs way better than the minimum divergence matching with respectively 7\% and 14\% failure of recovering the true matching.\\
Figure \ref{fig:barplot_n_clust_matching_comparison_trueC} shows the distribution of the estimated number of clusters in the full partition with the three matching algorithm for the three settings for clusters distribution across the shards and the two settings of clusters separability. The minimum divergence matching and the Ball matching allow to recover the true number of clusters in the vast majority of the cases. When the estimated number of clusters is not correct, the Ball matching tend to underestimate the true number of clusters by matching clusters that should not be matched whereas the Kullback-Leibler matching tends to overestimate this true number of clusters by failing to match clusters that should be matched.

\section{Fractionation of the prior} 
\label{app:prior_fractionation}
We perform a simulation study to compare in practice the behaviour of the posterior distribution of the partition of the data under the fractionated prior and the non-fractionated prior.\\
\newline
In order to assess the impact of fractionating or not the prior distribution, we compare the results in terms of estimated number of cluster and ARI in each of the local shards with and without prior fractionation. For this, we simulate a dataset of 50,000 individuals. We set the number of clusters to 20 and the probability of assignment in each of the clusters is drawn in a Dirichlet distribution where all the parameters are set to 0.5. This setting encourages unbalanced clusters. We define 10 categorical covariates - 4 with 2 modalities and 6 with 3 modalities -  following categorical distributions with cluster-specific parameters. For each cluster, the parameters of the categorical distributions on the covariates are drawn in a Dirichlet distribution where all the parameters are set to 0.1. In this setting, the parameters of the categorical distribution are close to the bounds of their definition domain (i.e. close to the values 0 or 1) and the clusters are really easily separable. \\
\newline
The individuals are then split randomly into balanced shards, we consider 5 cases varying the number of shards: 
\begin{itemize}
    \item 10 shards with 5000 individuals per shard; 
    \item 50 shards with 1000 individuals per shard; 
    \item 100 shards with 500 individuals per shard; 
    \item 250 shards with 200 individuals per shard; 
    \item 500 shards with 100 individuals per shard.
\end{itemize}
\begin{figure}
    \centering
    \includegraphics[width=0.8\linewidth]{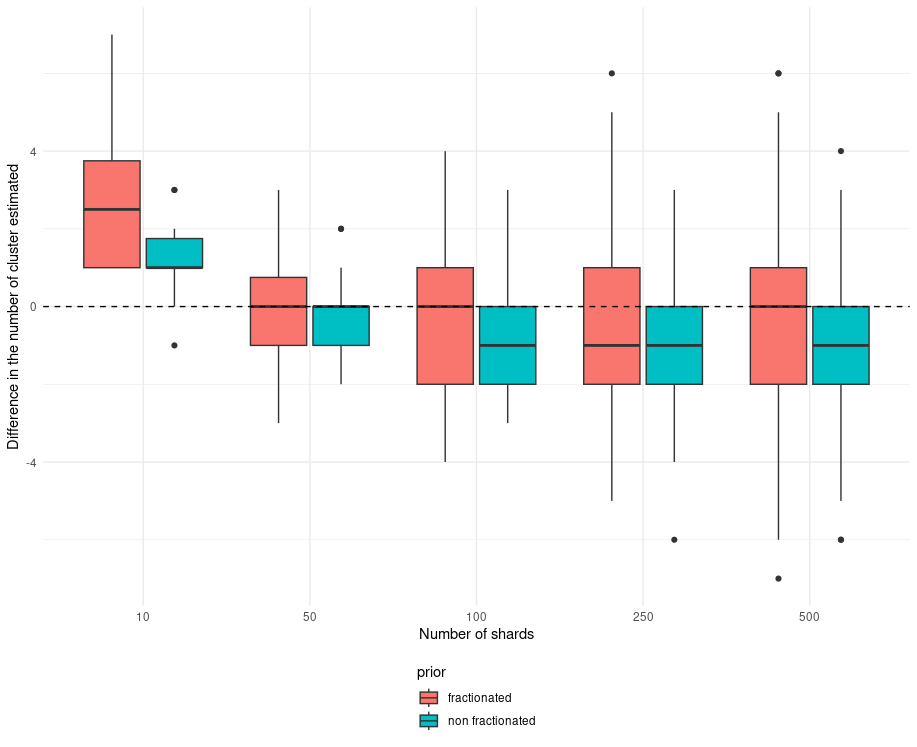}
    \caption{Difference between the number of estimated clusters and the true number of clusters in each shard for different number of shards.}
    \label{fig:diff_nb_clust_BL}
\end{figure}
Figure \ref{fig:diff_nb_clust_BL}  shows the difference between the number of estimated clusters and the true number of clusters in each shard for different number of shards. The impact of the prior fractionation on the error on  the number of estimated clusters depends deeply on the number of shards. When the number of shards is small, it can be assumed that all the clusters are observed in all the shards with sufficient information that they can be identified. In this case, with or without fractionation of the prior, the number of clusters is slightly overestimated without fractionation, but highly overestimated with fractionation. This might indicates that the MCMC is lacking observations to be able to empty all the extra mixture components. When the number of shards increases, some clusters become harder to identify because the number of individuals belonging to them in each shard is small. Without fractionation, the prior fails to allow enough uncertainty for the small clusters to be identified. In the case of a BL post-processing, fractionating the prior allows for more uncertainty in the number of estimated clusters with an error that is centred in zero. In comparison, the number of clusters tends to be underestimated without fractionation.\\
\newline 
\begin{figure}
    \centering
    \includegraphics[width=0.8\linewidth]{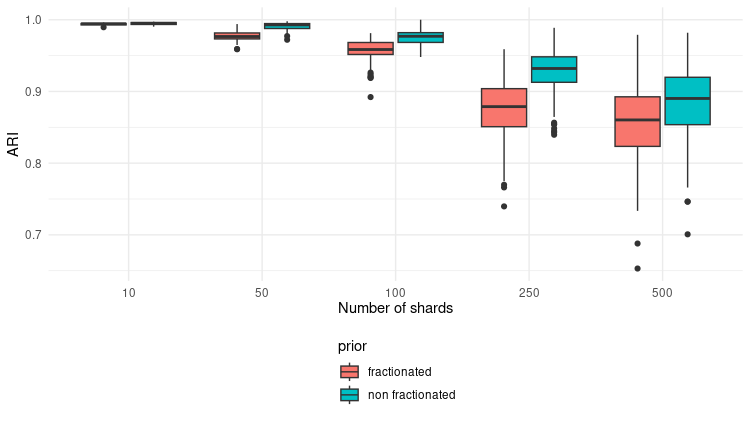}
    \caption{Distribution of the adjusted Rand index (ARI) between the estimated and the true local partitions in each shard for different number of shards.}
    \label{fig:ARI_prior_comparison}
\end{figure}
Figure \ref{fig:ARI_prior_comparison} shows the distributions of the adjusted Rand index between the estimated and the true local partitions for the considered numbers of shards. The local adjusted Rand index is on average higher without fractionation of the prior. This result can be explained by the fact that the metric is computed locally. Indeed, by doing so, we do not account for the fact that the data on which the metric is computed is actually a shard of a larger dataset. By ignoring this information at the comparison metric level, we do compare the clustering performances of a Bayesian mixture model with traditional prior distributions (the non-fractionated case) to the performance of a Bayesian mixture model with very much non standard prior distributions (the fractionated case). It would be interesting as a follow-up of this study to compare the fractionated prior in the full data after matching the local partitions. \\
\newline
To our knowledge, there exists no inference procedure to perform unsupervised clustering in a federated learning context that is able to split clusters at the global level. However, many of them are able to group clusters that comes from the same shard at the global level. For this reason, we do believe that overestimations of the number of clusters at the local level are easy to mitigate at the global level, but the errors induced by underestimating the number of clusters at the local level cannot be corrected at the global level. For this reason, we do believe that fractionating the prior at the local level is important.

\section{Post-processing the MCMC output}
\label{app:MCMC_postprocessing}
Posterior inference in mixture models via Markov Chain Monte Carlo (MCMC) algorithms does not straightforwardly yield a single, definitive clustering structure of the data. Because the MCMC output maps the marginal posterior distribution of all possible partitions, it offers the distinct advantage of quantifying clustering uncertainty. However, a post-processing step must be applied to extract a single representative partition from the ensemble.\\
\newline
A point-estimate partition can be identified by applying the Partitioning Around Medoids (PAM) algorithm to the posterior mean dissimilarity matrix. At any given MCMC iteration $t$, we define an $N \times N$ dissimilarity matrix whose $(i,j)$-th element equals 0 if individuals $i$ and $j$ reside in the same cluster, and 1 otherwise. Averaging these matrices across all post-burn-in iterations yields the mean dissimilarity matrix, where each element $(i,j)$ represents the posterior probability (or proportion of iterations) that individuals $i$ and $j$ are assigned to different clusters.\\
\newline
Once the single optimal partition $\hat{P}$ is identified, the MCMC chains of the model parameters must be realigned relative to this reference framework. Let $\hat{P}_i$ denote the cluster assignment of individual $i$ within $\hat{P}$, and let $\hat{\psi}_{c}^{(t)}$ represent the cluster-specific parameters for cluster $c$ at iteration $t$. We can derive the realigned parameter chains by mapping them through the individuals belonging to each target cluster: $$\hat{\psi}_{c}^{(t)} = \frac{\sum_{i = 1}^N \psi_{C_i^{(t)}}^{(t)} \mathbb{1}_{(\hat{P}_i = c)}}{\sum_{i = 1}^N \mathbb{1}_{(\hat{P}_i = c)}}$$
where $C_i^{(t)}$ is the cluster index of individual $i$ at iteration $t$, and $\psi_{C_i^{(t)}}^{(t)}$ is its corresponding parameter vector. \\
\newline
Similarly, the mixture weights $\hat{\pi}_c^{(t)}$ for the reference partition can be reconstructed by projecting the iteration-specific weights $\pi_{c'}^{(t)}$ based on the empirical overlap between the estimated clusters and the reference clusters: $$\hat{\pi}_c^{(t)} = \sum_{c'} \left( \frac{\sum_{i=1}^N \mathbb{1}_{(C_i^{(t)} = c') \cap (\hat{P}_i = c)}}{\sum_{i=1}^N \mathbb{1}_{(C_i^{(t)} = c')}} \right) \pi_{c'}^{(t)}.$$ This formulation mathematically guarantees that the realigned weights preserve their property of summing to one.

\section{Considered long term health condition in the THIN database}
\label{app:THIN}
Table \ref{tab:THIN_disease prevalence} shows the list of the 94 considered long term health condition and proportion of individuals suffering from each condition in the studied population. Hypertension is the most frequent condition in our population, affecting 63.2\% of the patients. It is then followed by ostheoarthisis (39.3\%), cataract (34.9\%), deafness (27.0\%), ischemic heart failure (26.8\%), arythmia (18.3\%) and artrial fibrilation (16.3\%). All the other considered diseases affect less than 15\% of the population with 60 of them affecting less than 5\% of the studied population and 32 of them affecting less than 1\% of the population. Polycystic ovary syndrome, eating disorder and human immunodeficiency virus  are the less frequent condition in our population, with less of 1\textperthousand \ patients affected.
\begin{table}[ht]
    \centering
    \resizebox{\textwidth}{!}{
    \begin{tabular}{|l|c||l|c|}
  \hline
Condition & Proportion & Condition & Proportion \\ 
  \hline
  Any cancer diagnosis & 0.138 & Endometriosis & 0.002 \\ 
  Lung cancer & 0.004 & Polycystic ovary syndrome & 0.000 \\ 
  Breast cancer & 0.038 & Pernicious anaemia & 0.020 \\ 
  Bowel cancer & 0.026 & Venous thromboembolism & 0.044 \\ 
  Prostate cancer & 0.028 & Pulmonary embolism& 0.023 \\ 
  Lymphoma & 0.005 & Coagulopathy & 0.001 \\ 
  Leukaemia & 0.005 & Depression & 0.136 \\ 
  Skin cancer & 0.126 & Anxiety& 0.094 \\ 
  Melanoma & 0.013 & Serious mental illness & 0.014 \\ 
  Metastatic cancer & 0.004 & Substance misuse & 0.006 \\ 
  Arrhythmia & 0.183 & Alcohol problem & 0.006 \\ 
  Atrial fibrillation & 0.163 & Attention deficit hyperactivity disorder & 0.006 \\ 
  Hypertension & 0.632 & Eating disorder & 0.000 \\ 
  Heart failure & 0.101 & Learning disability & 0.001 \\ 
  Ischemic heart disease & 0.268 & Alzheimer’s diseases & 0.048 \\ 
  Heart valve disease& 0.057 & Vascular dementia & 0.035 \\ 
  Cardiomyopathy & 0.003 & Dementia & 0.131 \\ 
  Congenital heart disease & 0.003 & Parkinson’s disease & 0.019 \\ 
  Peripheral vascular disease & 0.076 & Migraine & 0.037 \\ 
  Aortic aneurysm & 0.018 & Multiple sclerosis & 0.001 \\ 
  Transient ischaemic attack & 0.095 & Epilepsy & 0.017 \\ 
  Ischaemic stroke & 0.032 & Hemiplegia & 0.008 \\ 
  Haemorrhagic stroke & 0.007 & Chronic fatigue syndrome & 0.001 \\ 
  Stroke (not otherwise specified) & 0.104 & Fibromyalgia & 0.002 \\ 
  TIA stroke & 0.185 & Osteoarthritis & 0.393 \\ 
  Eczema & 0.160 & Osteoporosis & 0.144 \\ 
  Psoriasis & 0.040 & Polymyalgia & 0.050 \\ 
  Vitiligo & 0.002 & Rheumatoid arthritis & 0.023 \\ 
  Alopecia & 0.002 & Sj{\"o}gren’s syndrome & 0.002 \\ 
  Rhinoconjunctivitis & 0.068 & Systemic lupus erythematosus & 0.002 \\ 
  Sinusitis & 0.042 & Systemic sclerosis & 0.001 \\ 
  Deafness & 0.270 & Psoriatic arthritis & 0.002 \\ 
  Blindness & 0.059 & Ankylosing spondylitis & 0.002 \\ 
  Cataract & 0.349 & Gout & 0.076 \\ 
  Glaucoma & 0.098 & Chronic kidney disease & 0.332 \\ 
  Age-related macular degeneration & 0.085 & Asthma & 0.122 \\ 
  Retinopathy & 0.080 & Chronic obstructive pulmonary disease & 0.082 \\ 
  Uveitis & 0.015 & Obstructive sleep apnoea & 0.002 \\ 
  Scleritis & 0.002 & Bronchiectasis & 0.010 \\ 
  Peptic ulcer & 0.069 & Pulmonary fibrosis & 0.005 \\ 
  Inflammatory bowel disease & 0.011 & Hyperthyroidism & 0.027 \\ 
  Irritable bowel syndrome & 0.047 & Hypothyroidism & 0.131 \\ 
  Liver disease & 0.004 &  Type 1 diabetes mellitus & 0.006 \\ 
  Nonalcoholic steatohepatitis/Nonalcoholic fatty liver & 0.001 &  Type 2 diabetes mellitus& 0.147 \\ 
  Diverticulitis & 0.146 & Human immunodeficiency virus& 0.000 \\ 
  Coeliac disease & 0.003 & Benign prostatic hyperplasia & 0.094 \\ 
  Pancreatitis & 0.001 & Erectile dysfunction & 0.021 \\ 
   \hline
\end{tabular}
}
    \caption{List of the considered long term health condition and proportion of individuals suffering from each condition in the studied population}
    \label{tab:THIN_disease prevalence}
\end{table}

\section{Supplementary results on THIN}
\label{app:THIN_cluster}
\begin{figure}
    \centering
    \includegraphics[width=1\linewidth]{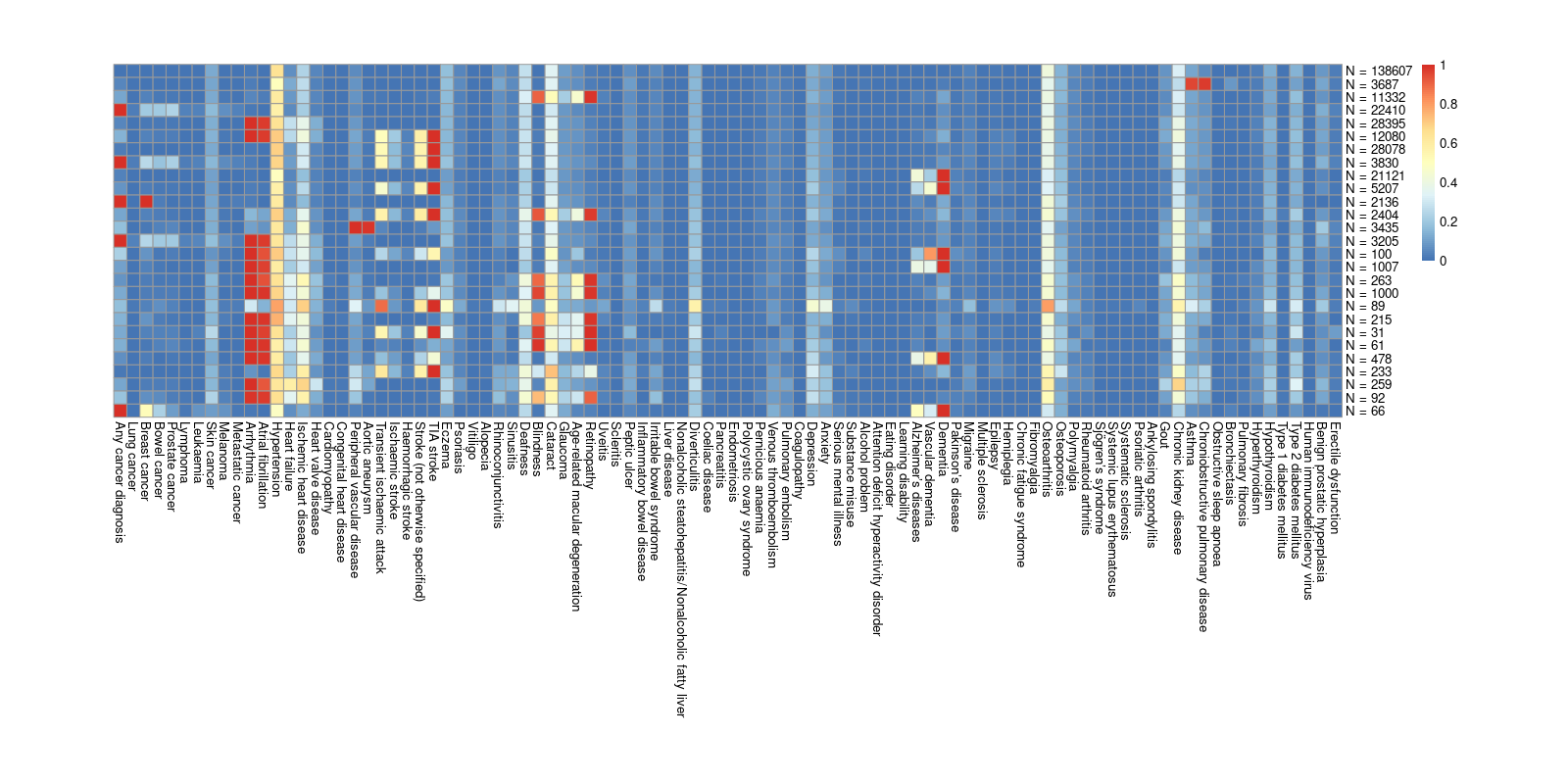}
    \caption{Proportion of individuals presenting each disease in each cluster}
    \label{fig:THIN_cluster_summary_appendix}
\end{figure}

Figure \ref{fig:THIN_cluster_summary_appendix} shows the proportion of individuals presenting each disease in each cluster. As most of the considered conditions have a very low prevalence in the considered population, they appear to have a low prevalence in all the clusters. However, some clusters stand out for being formed almost exclusively formed of patients suffering from a specific disease, as cluster 2 for asthma and chronic-obstructive pulmonary disease, cluster 11 with breast cancer or cluster 13 with congenital heart disease and aortic aneurysm. \\
Table \ref{tab:THIN_supp_res_clusters_characterisation} shows the proportion of mortality, the mean age at death and the proportion of female per cluster in the considered THIN subset. Cluster 11 is composed at 99\% of females. As mentioned above, this cluster is composed of individuals suffering of breast cancer, which is a disease affecting male in only less than 1\% of the cases in the UK according to cancer research UK. Even though, there exists some statistically significant differences in the mean age at death and percentage of mortality across the clusters, those differences are not big enough in practice to deliver a sensible interpretation.
\newline

\begin{table}[ht]
    \centering
    \resizebox{\textwidth}{!}{
    \begin{tabular}{|c||c|c||c|c||c|c|}
    \hline
        Cluster & Mortality (\%) & p-value & Age at death (Mean) & p-value & Sex (\% of females) & p-value \\
    \hline
         1  & 0.40 & $<$0.0001 & 89.65 & $<$0.0001  & 0.67 & $<$0.0001 \\
         2  & 0.56 & $<$0.0001 & 87.53 & $<$0.0001 & 0.55 & $<$0.0001 \\
         3  & 0.52 & $<$0.0001 & 91.18 & $<$0.0001 & 0.71 & $<$0.0001 \\
         4  & 0.49 & 0.0007 & 88.61 & $<$0.0001 & 0.53 & $<$0.0001 \\
         5  & 0.55 & $<$0.0001 & 89.24 & $ <$0.0001 & 0.59 & $<$0.0001 \\
         6  & 0.61 & $<$0.0001 & 89.11 & $<$0.0001 & 0.58 & $<$0.0001 \\
         7  & 0.51 & $<$0.0001 & 89.40 & 0.6114 & 0.61 & $<$0.0001 \\
         8  & 0.57 & $<$0.0001 & 88.72 & $<$0.0001 & 0.53 & $<$0.0001 \\
         9  & 0.61 & $<$0.0001 & 89.42 & 0.9314 & 0.77 & $<$0.0001 \\
         10 & 0.66 & $<$0.0001 & 89.11 & 0.0010 & 0.71 & $<$0.0001 \\
         11 & 0.44 & $<$0.0001 & 89.33 & 0.5729 & 0.99 & $<$0.0001 \\
         12 & 0.56 & $<$0.0001 & 90.65 & $<$0.0001 & 0.69 & $<$0.0001 \\
         13 & 0.61 & $<$0.0001 & 87.88 & $<$0.0001 & 0.31 & $<$0.0001 \\
         14 & 0.61 & $<$0.0001 & 88.63 & $<$0.0001 & 0.52 & $<$0.0001 \\
         15 & 0.65 & 0.0010 & 88.51 & 0.1187 & 0.67 & 0.7006 \\
         16 & 0.68 & $<$0.0001 & 89.27 & 0.3921 & 0.72 & $<$0.0001 \\
         17 & 0.61 & $<$0.0001 & 91.10 & $<$0.0001 & 0.63 & 0.6463 \\
         18 & 0.63 & $<$0.0001 & 90.65 & $<$0.0001 & 0.63 & 0.2069 \\
         19 & 0.61 & 0.0229 & 87.78 & 0.0031 & 0.70 & 0.3810 \\
         20 & 0.57 & 0.0132 & 90.75 & 0.0016 & 0.70 & 0.1351 \\
         21 & 0.55 & 0.5649 & 90.88 & 0.2569 & 0.61 & 0.8378 \\
         22 & 0.59 & 0.1131 & 91.97 & 0.0570 & 0.72 & 0.2773 \\
         23 & 0.64 & $<$0.0001 & 88.83 & 0.0264 & 0.64 & 0.6615 \\
         24 & 0.64 & $<$0.0001 & 89.49 & 0.8693 & 0.69 & 0.1777 \\
         25 & 0.59 & 0.0005 & 88.69 & 0.0380 & 0.52 & $<$0.0001\\
         26 & 0.62 & 0.0104 & 90.16 & 0.2202 & 0.63 & 0.8294 \\
         27 & 0.67 & 0.0037 & 88.68 & 0.2577 & 0.79 & 0.0231 \\
         \hline
    \end{tabular}
    }
    \caption{Proportion of mortality, mean age at death and proportion of female per cluster in the considered THIN subset.}
    \label{tab:THIN_supp_res_clusters_characterisation}
    \flushleft
    \underline{Note 1 :} P-values are derived from z-test for proportion equality in the clusters versus in the complete studied population with continuity correction for mortality and sex, and from Welch's t-test for means equality in the clusters versus in the complete studied population for age at death. \\
    \underline{Note 2 :} The value of the Bonferroni threshold leading to a family-wise error rate less than or equal to 5\% for 81 statistical tests is 0.0006.
\end{table}




\end{document}